%% file: main.tex
\documentclass[10pt,twocolumn,letterpaper]{article}

\usepackage{cvpr}              

\usepackage{amsfonts}       
\usepackage{nicefrac}       
\usepackage{microtype}      
\usepackage{xcolor}         
\usepackage{bm}
\usepackage{amsmath}
\usepackage{graphicx}
\usepackage{algorithm}
\usepackage{algorithmic}
\usepackage{multirow}
\usepackage{amsthm}


\definecolor{cvprblue}{rgb}{0.21,0.49,0.74}
\usepackage[pagebackref,breaklinks,colorlinks,allcolors=cvprblue]{hyperref}

\def\paperID{ 9952 } 
\def\confName{CVPR} 
\def\confYear{2025} 

\title{Deterministic-to-Stochastic Diverse Latent Feature Mapping for Human Motion Synthesis}

\author{
\small Yu Hua \\
\small Nanyang Technological University \\
\small yu\_hua@ntu.edu.sg
\and
\small Weiming Liu\\
\small ByteDance Inc.\\
\small lwming95@gmail.com
\and
\small Gui Xu\\
\small Dalian University\\
\small guixu@s.dlu.edu.cn
\and
\small Yaqing Hou\\
\small Dalian University of Technology\\
\small houyq@dlut.edu.cn
\and
\small Yew-Soon Ong\\
\small Nanyang Technological University\\
\small CFAI, A*STAR\\
\small asyong@ntu.edu.sg
\and
\small Qiang Zhang\\
\small Dalian University of Technology\\
\small zhangq@dlut.edu.cn
}

\begin{document}
\maketitle

\setlength{\floatsep}{4pt plus 4pt minus 1pt}
\setlength{\textfloatsep}{4pt plus 2pt minus 2pt}
\setlength{\intextsep}{4pt plus 2pt minus 2pt}
\setlength{\dbltextfloatsep}{3pt plus 2pt minus 1pt}
\setlength{\dblfloatsep}{3pt plus 2pt minus 1pt}
\setlength{\abovecaptionskip}{3pt}
\setlength{\belowcaptionskip}{2pt}
\setlength{\abovedisplayskip}{2pt plus 1pt minus 1pt}
\setlength{\belowdisplayskip}{2pt plus 1pt minus 1pt}

\input{sec/0Abstract}    
\input{sec/1Intro}

\input{sec/2RelatedWork}
\input{sec/3Preliminariy}

\input{sec/4Method}
\input{sec/5Experiment}
\input{sec/6Conclusion}

\section{Acknowledgment}

This research is supported by the National Research Foundation, Singapore under its AI Singapore Programme (AISG Award No: AISG3-RP-2022-031), the National Research Foundation, Singapore and DSO National Laboratories under the AI Singapore Programme (AISG Award No.: AISG2-GC-2023-010, ``Design Beyond What You Know": Material-Informed Differential Generative AI (MIDGAI) for Light-Weight High-Entropy Alloys and Multi-functional Composites (Stage 1a), the National Natural Science Foundation of China under Grant 62372081, the Young Elite Scientists Sponsorship Program by CAST under Grant 2022QNRC001, the Liaoning Provincial Natural Science Foundation Program under Grant 2024010785-JH3\/107, the Dalian Science and Technology Innovation Fund under Grant 2024JJ12GX020, the Dalian Major Projects of Basic Research under Grant 2024JJ12GX020 2023JJ11CG002 and the 111 Project under Grant D23006.

{
    \small
    \bibliographystyle{ieeenat_fullname}
    \bibliography{main}
}

\input{sec/appendix}

\end{document}

%% file: sec/0Abstract.tex
\begin{abstract}
Human motion synthesis aims to generate plausible human motion sequences, which has raised widespread attention in computer animation. 
Recent score-based generative models (SGMs) have demonstrated impressive results on this task. However, their training process involves complex curvature trajectories, leading to unstable training process.
In this paper, we propose a Deterministic-to-Stochastic Diverse Latent Feature Mapping (DSDFM) method for human motion synthesis.
DSDFM consists of two stages. The first human motion reconstruction stage aims to learn the latent space distribution of human motions. 
The second diverse motion generation stage aims to build connections between the Gaussian distribution and the latent space distribution of human motions, thereby enhancing the diversity and accuracy of the generated human motions. This stage is achieved by the designed deterministic feature mapping procedure with DerODE and stochastic diverse output generation procedure with DivSDE. 
DSDFM is easy to train compared to previous SGMs-based methods and can enhance diversity without introducing additional training parameters.
Through qualitative and quantitative experiments, DSDFM achieves state-of-the-art results surpassing the latest methods, validating its superiority in human motion synthesis.
\end{abstract}

%% file: sec/1Intro.tex

\section{Introduction}
\begin{figure*}
    \centering
    \includegraphics[width=0.78\linewidth]{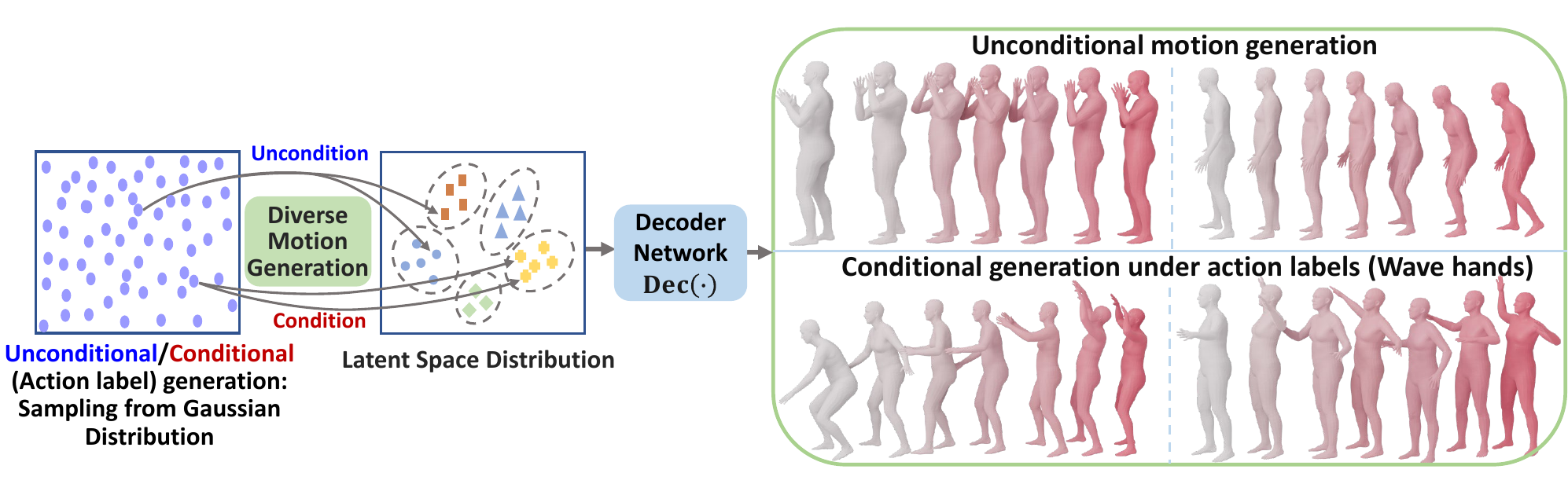}
    \caption{Examples of the inference process for human motion synthesis. Our method aims to generate diverse and accurate human motion sequences through the designed generative model. }
    \label{fig:first} 
\end{figure*}

Human motion synthesis task aims to generate diverse and high quality 3D human motion sequences. This task has wide-ranging applications, such as human motion understanding \cite{diomataris2024wandr, jin2024act, NEURIPS2023_4f8e27f6}, human-robot interactions \cite{wang2024incomplete, zhang2024motiongpt}, and computer graphics \cite{tang2024temporal}. 
Recent efforts mainly focus on conditional and unconditional human motion generation. Conditional human motion generation aims to generate human motion sequences under some limiting factors, such as music \cite{li2021learn, li2021ai}, audio \cite{alexanderson2023listen, liu2022audio, li2021audio2gestures}, and action labels \cite{petrovich2022temos, zhang2024motiongpt, petrovich2022temos}.
Unconditional human motion generation intends to generate diverse human motions \cite{petrovich2021action, raab2023modi} from diverse data, which still presents a significant challenge, especially when the human motion datasets are unstructured. 
In this paper, we focus on conditional (under the action labels) and unconditional human motion generations, as shown in Figure~\ref{fig:first}.
Efficiently generating diverse and accurate human motions remains a tremendous challenge, which has led to the development of many different generative models.

Recent advancements in deep generative models, including Variational Autoencoders (VAEs) \cite{yan2018mt, yuan2020dlow, zhang2021we}, Generative Adversarial Networks (GANs) \cite{odena2017conditional}, score-based generative models (SGMs), and related techniques \cite{ho2020denoising, wei2023human, yuan2023physdiff, song_score-based_2021, vahdat2021score}, emerge as the dominant approaches for capturing the data distribution. 
Specifically, VAEs leverage an encoder-decoder
network to learn the latent representation of human motion distribution. VAEs require approximate variational or Monte Carlo inference techniques, which tend to be intractable for complex models. 
%
GANs utilize a generator and discriminator to generate real-like motions from random noise. Unfortunately, GANs are known to suffer from numerical instability and mode collapse issues.
SGMs define a forward diffusion process that maps data to noise by gradually perturbing the input data. Generation corresponds to a reverse process that synthesizes novel data via iterative denoising process.
Even though they have presented high fidelity in generation, it is important to note that these methods have the challenge of curve trajectory modeling within diffusion models, as their forward pass is inherently designed to exhibit curvature in SDE, following either a Variance Preserving SDE (VPSDE) or a Variance Exploding SDE (VESDE) \cite{song_score-based_2021}, leading to unstable training process and slow inference process.
Recent methods, like DDIM \cite{song2020denoising}, aim to accelerate the inference process by one-step or few-step generator, nevertheless, these methods lead to an obvious performance drop \cite{davtyan2023efficient}, and the training process is still unstable.

To synthesize diverse and accurate human motions, we propose a novel method called DSDFM for human motion synthesis. 
The proposed method has straight trajectories and is easy to train compared to previous SGMs methods, while guaranteeing the diversity and accuracy of the generated human motions.
The proposed DSDFM consists of two stages. 
In the first stage, a human motion reconstruction process is designed to learn the latent space distribution of human motions and motion representation. This process is implemented by the Vector Quantized Variational Autoencoders (VQVAE) \cite{van2017neural} network. 
In the second stage, we design a diverse motion generation module, including deterministic feature mapping procedure and stochastic diverse output generation procedure.  
Deterministic feature mapping procedure aims to explore the optimal solution for building the connections between the Gaussian distribution and the latent space distribution of human motions using the designed Deterministic Ordinary Equation (DerODE) operation. DerODE has a straight training trajectory compared to previous diffusion generative methods \cite{ho2020denoising, song_score-based_2021} and Flow Matching \cite{liu2022flow}.
Although DerODE is easy to train, it is hard to generate highly diverse human motion patterns since DerODE could only provide deterministic output.
Therefore, the designed stochastic diverse output generation procedure aims to enhance the diversity of generated human motions through Diverse Stochastic Differential Equations (DivSDE). It is noted that DivSDE operates during the sampling process of the model without introducing additional training processes.

In summary, our main contributions are as follows: 
\begin{itemize}
    \item We propose a novel method called Diverse Latent Feature Mapping (DSDFM) for human motion synthesis. DSDFM is efficient to train and to utilize at sampling process, and can be used for conditional and unconditional generation. 
    \item We propose an optimal solution to build the connection between the Gaussian distribution and the latent space distribution of human motions. In addition, we provide a stochastic diverse output generation process during the sampling process without reintroducing additional training processes.
    \item The proposed method DSDFM is evaluated on widely-used human motion datasets in the comprehensive experiments. The obtained results demonstrate the effectiveness of the proposed method over the state-of-the-art approaches for conditional and unconditional human motion generation tasks.
\end{itemize}

%% file: sec/2RelatedWork.tex
\begin{figure*}
    \centering
    \includegraphics[width=0.9\linewidth]{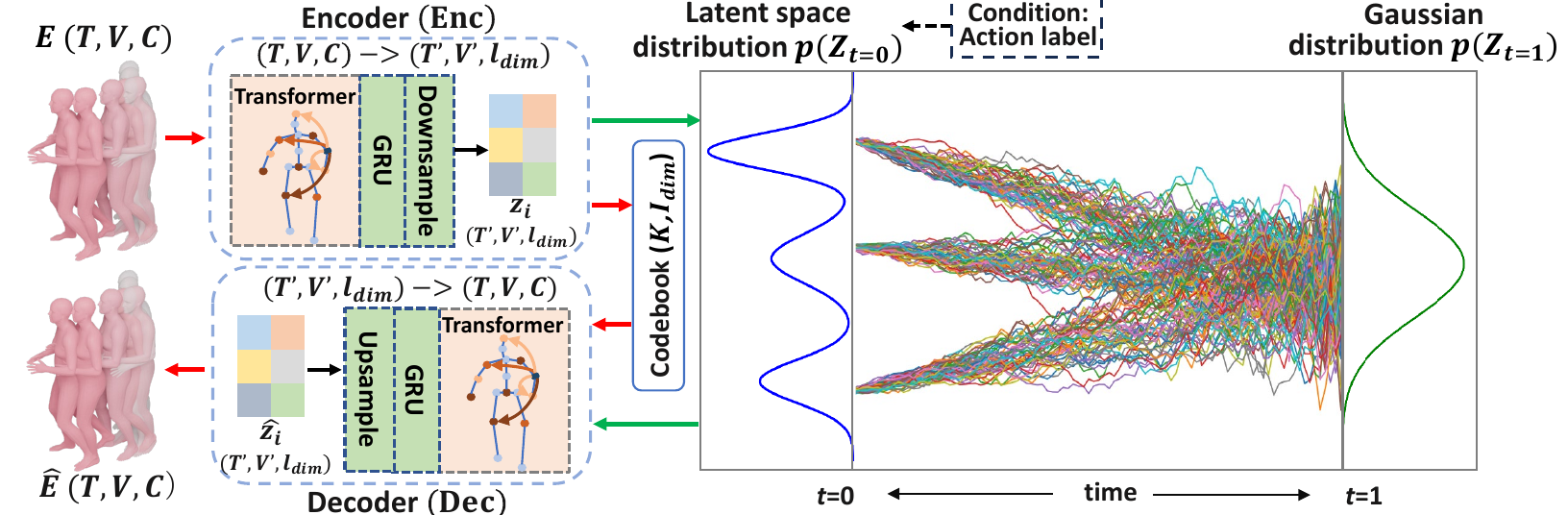} 
    \caption{The overview of the proposed method DSDFM. The red arrow denotes the first stage and the green arrow denotes the second stage of DSDFM. }
    \label{fig:overview}
\end{figure*}

\section{Related Work}
\subsection{Human Motion Synthesis}
Conditional human motion synthesis aims to generate diverse and realistic human motions \cite{NEURIPS2023_4c9477b9, NEURIPS2022_6030db51, yu1, yu3} according to various conditional inputs, such as action labels \cite{chen2023executing, sun2021action, Zhang_2023_CVPR} and music \cite{guo2024momask, 10.1145/3610543.3626164}. 
For example, MDM \cite{tevet_human_2022} utilized a diffusion-based generative model for action-conditioned human motion generations, and reported a trading-off between diversity and fidelity of human motions due to the curve trajectory of training and sampling process. 
MLD \cite{chen2023executing} proposed to utilize the DDPM in latent space for human motion generations given an input action label, which also encountered the same problem as DDPM. 
In addition, the unconditional human motion synthesis \cite{raab2023modi, cervantes2022implicit, yu2} task also encounters the same issues although a series of achievements have been made in this field. For example, Holden et al. \cite{holden2016deep} presented a pioneer work in deep unconditional human motion synthesis. Modi \cite{raab2023modi} employed the style of StyleGAN to synthesize human motions. Unfortunately, these methods usually suffer from mode collapse or mode mixture.  
In contrast, we propose a novel method for conditional and unconditional human motion synthesis, which is easy to train compared to previous diffusion-based methods while guaranteeing the diversity of generated motions.

\subsection{ Diffusion Generative Models}
Recent years have witnessed a promising potential in modeling data distributions with diffusion generative models using Denoising diffusion probabilistic modeling (DDPM) \cite{ho2020denoising} and score-based generative models (SGMs) \cite{song_score-based_2021}, which define a forward diffusion process that maps data to noise by gradually perturbing the input data.
Variants of SGMs and techniques have been applied to images \cite{dhariwal2021diffusion}, audio \cite{mittal2021symbolicdiffusion}. 
For example, Robin et al. \cite{gao2020learning} proposed latent diffusion models (LDMs) that work on a compressed latent space of lower dimensionality for high-resolution image synthesis.
LSGM \cite{vahdat2021score} proposed to train SGMs in a latent space, which relies on the variational autoencoder framework to generate diverse images.
However, it is important to note that these methods have the challenge of curve trajectory modeling within diffusion models, as their complex forward and backward processes are inherently designed to exhibit curvature, leading to unstable training process and slow sampling process.
Although DDIM and related techniques \cite{song2020denoising} can shorten the sampling process, they often result in a performance drop \cite{davtyan2023efficient,wu2023fast}.
Flow Matching-based methods \cite{lipman2023flow, liu2022flow} offer a more robust and stable alternative to diffusion models during the training process. However, the trajectories between source and target distributions remain relatively curved, and more importantly, these methods struggle to produce highly diverse samples, often sacrificing diversity in the training process. 
In contrast, we propose a generative model DSDFM for human motion generation tasks. This model utilizes straight trajectories, making it easier to train compared to other diffusion models. Additionally, it is capable of generating diverse human motion sequences.



%% file: sec/3Preliminariy.tex
\section{Preliminary}
Score-based diffusion models gradually perturb data by a forward diffusion process, and
then reverse it to recover the data \cite{song2020improved, song_score-based_2021}.
Under the stochastic differential equation (SDE) framework proposed in \cite{song_score-based_2021}, diffusion models construct a process ${x(t)}^T_{t=0}$ indexed by a continuous time variable $t \in [0, T]$, such that $x(0)\sim p_0$, 
for which we have a dataset of i.i.d. samples, and $x(T) \sim p_T$, we have a tractable form to generate samples efficiently. $p_0$ is the data distribution, $p_T$ is the prior distribution. The forward diffusion process can be modeled as the solution to an $\mathrm{It}\hat{0}$ SDE:
\begin{equation}
    dx_t = f(x,t)dt+g(t)dw_t,
\end{equation}
where $w$ is the standard Wiener process (a.k.a., Brownian motion), $f(\cdot,t):\mathbb{R}^d\to\mathbb{R}^d$ is a vector valued function called the $drift$ coefficient of $x(t)$, and $g(\cdot): \mathbb{R}\to\mathbb{R}$ is a scalar function known as the diffusion coefficient of $x(t)$. There are various ways of designing the SDE such that it diffuses the data distribution into a fixed prior distribution $p_T$. 
By starting from samples of $x(T) \sim p_T$ and reversing the process, we can obtain samples $x(0) \sim p_0$. The reverse of a diffusion process is also a diffusion process, running backwards in time and given by the reverse-time SDE:
\begin{equation}
    dx_t=[ f(x,t)- g(t)^2 \nabla_{x} \log p_t(x)] dt+ g(t) d\bar{w}_t,
\end{equation}
where $\bar{w}$ is a standard Wiener process when time flows backwards from $T$ to 0, $dt$ is an
infinitesimal negative timestep. Once the score $\nabla_{x}\log p_t(x)$ is learned, we can derive the reverse diffusion process and simulate it to sample from $p_0$.


%% file: sec/4Method.tex
\section{The Proposed Method}
This paper aims to synthesize diverse and realistic human motion sequences. The overview of the proposed method is shown in Figure \ref{fig:overview}. 
The conditional motion generation is performed under the action labels (Action-to-Motion task). Once the action labels are removed, the entire process becomes unconditional motion generation.
The training process of the DSDFM mainly involved in two stages. The first stage is the human motion reconstruction process (Section 4.1), which aims to learn the human motion representation and capture the latent space distribution of human motions. The second stage (Section 4.2) aims to build the connections between the Gaussian distribution and the latent space distribution using the designed deterministic feature mapping procedure (DerODE) (Section 4.2.1).
Moreover, we employ the stochastic diverse generation process (DivSDE) to enhance the diversity of generated human motions (Section 4.2.2).

\subsection{Human Motion Reconstruction}
The human motion reconstruction network aims to learn the representation and the latent space distribution of human motions. In this process, we utilize VQVAE \cite{van2017neural} to capture dynamic spatio-temporal features of human motions.
Specifically, the input is a sequence of human motion sequence $\bm{E}=\{ e_1, e_2, \cdots, e_T \}$ with the length of $T$, where $ e_t \in \mathbb{R}^{V \times C} $ is denoted by 3D coordinates at time $t$, $C$ denotes the 3D coordinates of human joints ($C=3$), $V$ is the used number of human joints. The encoder of VQVAE aims to transform motion sequence into latent features, i.e., ${\rm Enc}(\bm{E}) \rightarrow z_i \in \bm{Z}$. $z_i$ is substituted by its closest vector $k_j$ using a quantization codebook, where $\hat{z}_i=\arg\min_{ k_j\in K }|| z_i - k_j ||$. The quantized feature $\hat{z}_i$ is decoded to $\hat{\bm{E}}$ by the decoder network, i.e., ${\rm Dec}(\hat{z}_i) \rightarrow \bm{\hat{E}}$.

In this work, the encoder ${\rm Enc}(\cdot)$ and decoder ${\rm Dec}(\cdot)$ networks are implemented by the Transformer \cite{vaswani2017attention} and GRU \cite{chung2014empirical} module. 
For the Transformer process, we project the input human motion sequences into matrices ${Q}$, ${K}$, and ${V}$ by ${W}_Q,{W}_K,{W}_V$. 
The summary of the spatial joints $ {\tilde{M}}_{t}$ is calculated by aggregating all the joint information using the multi-head mechanism (${\rm head}_i$). 
The $GRU_\phi$ with parameter $\phi$ intends to capture the smoothness property of human motions, and then encode the human motions into latent space $\bm{Z}$.
%
In addition, the decoder ${\rm Dec}(\cdot)$ aims to map the latent space $\bm{Z}$ back to the reconstructed human motion sequence. 
The VQVAE is optimized by minimizing the following loss function:
\begin{equation}
\begin{aligned}
\mathcal{L}_{VQ}=& \mathcal{L}( {\bm{E}},\hat{\bm{E}}) 
+||\hat{z}-\operatorname{sg}(z)||_{2}^{2}+\beta||\operatorname{sg}(\hat{z})-z||_{2}^{2},
\end{aligned}
\end{equation}
where $\mathrm{sg}[\cdot]$ is the stop gradient operator and $\beta$ is the hyper parameter. The first term $\mathcal{L}(\bm{E},\hat{\bm{E}})=\sum_{t=1}^{T}\sum_{v=1}^{V}\parallel e^{(v)}_t-\hat{e}^{(v)}_t\parallel_2 $, represents the reconstruction error. The second term aims to optimize the codebook, and the last term is to optimize the encoder by pushing $z$ close to its nearest latent vector in the codebook.
The human motion reconstruction process aims to learn the human motion representation and map the human motions into latent space $\bm{Z}$.

\subsection{Diverse Motion Generation}
Although we have established the human motion reconstruction in Section 4.1, we still cannot generate diverse human motion accordingly.
The main reason is that the latent feature space for human motion is rather complicated and hard to sample.
Therefore, it is essential to model the latent feature space for human motion by establishing the relationship between a Gaussian distribution and the latent space distribution.
Previous diffusion-based generative methods \cite{song_score-based_2021} and flow matching methods \cite{lipman2023flow}, suffer from instability during training, exhibiting curved trajectories or difficulty in generating diverse samples.
To tackle this issue, we innovatively propose a diverse motion generation module to enhance the diversity and accuracy of generated human motion sequences.
Diverse motion generation module consists of two steps, i.e., \textit{deterministic feature mapping procedure} and \textit{stochastic diverse output generation procedure}. 
We will introduce the details of our proposed diverse motion generation module in this section.

\subsubsection{Deterministic Feature Mapping Procedure}
To start with, we first introduce the deterministic feature mapping procedure.
The deterministic feature mapping procedure is designed to model the relationship between Gaussian distribution $p(\bm{Z}_{t=1})$ and the latent distribution for human motion $p(\bm{Z}_{t=0})$ efficiently.
Specifically, we propose Deterministic Ordinary Equation (DerODE) operation in the deterministic feature mapping procedure by depicting the transformation with Proposition 1 to achieve the corresponding goal.
%


%
\textbf{Proposition 1.} \textit{Given the ordinary equation $d\bm{z}_t = u(\bm{z}_t, t)dt$, where $u(\bm{z}_t, t)$ denotes the drift function, and suppose the probability of data distribution $z(t)$ is set to be $p(z(t)) = \mathcal{N}(\mu(t),\sigma^{2}(t))$ at the time step $t$, where $\mu(t)$ and $\sigma(t)$ denote the mean and variance of the Gaussian distribution respectively, the drift function $u(\bm{z}_t, t)$ can be shown as:  }
\begin{equation}
   u(z_t,t)=\sigma^{\prime}(t)\cdot\frac{z(t)-\mu(t)}{\sigma(t)}+\mu^{\prime}(t).
\end{equation}

The illustrations of Proposition 1 can be found in \cite{lipman2022flow}. 
We can utilize Proposition 1 to transform the data across different distributions.
Specifically, we need to establish the connections among the latent motion feature space $p(\bm{Z}_{t=0})$ and the standard Normal distribution $p(\bm{Z}_{t=1}) = \mathcal{N}(0, \bm{I})$ by carefully designing $\mu(t)$ and $\sigma(t)$ for the downstream generation task.
However, previous methods (e.g., Flow Matching \cite{lipman2022flow}) just randomly sample data across different distributions, leading to less efficient model training and inference.  
To get straighter paths for the training process, we can introduce the optimal transport (OT) theory into this task. As discussed in \cite{peyre2019computational,liu2022exploiting,liu2024reducing,liu2024learning,liu2023joint,liu2023contrastive,liu2024user}, the OT problem aims to minimize the displacement cost between two distributions. Thus, we can leverage the transport plan $\boldsymbol{\pi}$ to build connections between two different distributions.
The calculation of the optimal transport $\boldsymbol{\pi}$ can be formulated as:
\begin{equation} 
\begin{aligned}
\left.
\begin{aligned}
&\min_{\boldsymbol{\pi} \in \Delta} J_{\rm OT} = \langle \boldsymbol{\pi}, \boldsymbol{C} \rangle  \\&s.t.\,\, \Delta = \left\{ \sum_{j=1}^N \pi_{ij} = a_i,\quad \sum_{i=1}^N \pi_{ij} = b_j, \quad \pi_{ij} \ge 0\right\},
\end{aligned}
\right.  
\end{aligned}
\end{equation} 
where $\Delta$ denotes the constraints on $\boldsymbol{\pi}$. 
%
%
$\bm{C}$ denotes the cost distance matrix which can be calculated as $C_{ij} = ||\bm{z}_{0,i} - \bm{z}_{1,j}  ||_2^2$ accordingly, where $\bm{z}_{0,i} \sim p(\bm{Z}_{t=0})$ and $\bm{z}_{1,j} \sim p(\bm{Z}_{t=1})$.
The optimization process for solving $\bm{\pi}$ has been provided in the Appendix \ref{sec:p1}.
%
%
%
Then we can obtain the matched data samples $(\bm{z}_{0,i}, \bm{z}_{1,j}) \sim \bm{\pi}$ via the coupling matrix.
Hence we can utilize the dynamic process $p(\bm{z}, t)$ on $\bm{\mu}(t) = t\bm{z}_{0,i} + (1-t)\bm{z}_{1,j}$ and $\bm{\sigma}(t) = \bm{0}$ where $t \in [0,1]$ as:
\begin{equation}  
p(\bm{z}_t, t) = \mathcal{N}(t\bm{z}_{1,j} + (1-t)\bm{z}_{0,i},  \bm{0}).
\end{equation}

Meanwhile we can obtain the drift function $\bm{u}(\bm{z}, t)$ via using the Proposition 1 as below: 
\begin{equation}  
\bm{u}(\bm{z}_t, t) =  \bm{\mu}'(t ) + \frac{\bm{z}(t) - \bm{\mu}(t)}{\bm{\sigma}(t)} \bm{\sigma}'(t ) =  \bm{z}_{1,j}  - \bm{z}_{0,i}.
\end{equation}

Specifically, we can employ a neural network $v_{\theta}(\cdot)$ with matching samples $(\bm{z}_0, \bm{z}_1) \sim \bm{\pi}$ to predict the deterministic drift $u(x,t)$ using the drift-estimate loss function:  
\begin{equation}   
\label{equ:62}
\begin{aligned}
&\min_{\boldsymbol{\theta}} J_{\rm drift} = \mathbb{E}_{(\boldsymbol{z}_0,\boldsymbol{z}_1) \sim \bm{\pi}} \left[ ||\bm{v}_{\theta}(\boldsymbol{z}_t, t) - (\boldsymbol{z}_1 - \boldsymbol{z}_0) ||_2^2 \right] .
\end{aligned} 
\end{equation} 

Moreover, we intend to figure out more consistent results \cite{song2023consistency} for achieving better performance.
That is, the coupling data samples with different time interpolation should have the same drift output 
as expected.
Therefore, we propose drift-consistent loss function:
\begin{equation}  
\label{equ:62}
\begin{aligned}
&\min_{\boldsymbol{\theta}} J_{\rm CL} = \mathbb{E}_{t,t'\in U[0,1]} \left[ ||\bm{v}_{\theta}(\boldsymbol{z}_t, t) - \bm{v}_{\theta}(\boldsymbol{z}_{t’}, t') ||_2^2 \right], \\
&{\rm where}\,\, \boldsymbol{z}_{t} = (1-t)\boldsymbol{z}_0 + t\boldsymbol{z}_1, \quad \boldsymbol{z}_{t'} = (1-t')\boldsymbol{z}_0 + t'\boldsymbol{z}_1
\end{aligned} 
\end{equation}

Finally, we combine the drift-estimate and drift-consistent loss functions for training our proposed DerODE:
\begin{equation}  
\label{equ:62}
\begin{aligned}
&\min_{\boldsymbol{\theta}} J_{\rm DerODE} = J_{\rm drift} + \lambda_{cl} J_{\rm CL},
\end{aligned} 
\end{equation}
where $\lambda_{cl}$ denotes the balanced parameter.
It is noticeable that DerODE will not involve complex denoising or score estimation procedures during the training stage.
Therefore, it could be much easier to train compared with other diffusion approaches.
Once we obtain the optimal solution on $\bm{v}^*(\cdot)$, we can generate new motion features in the latent space via randomly sample noise in the standard Gaussian distribution via:
\begin{align}  
& \widetilde{\bm{z}}_{0,i} = \widetilde{\bm{z}}_{1,i} -  \boldsymbol{v}_{\theta}(\widetilde{\bm{z}}_{1,i}, t=1) = {\rm DerODE}(\widetilde{\bm{z}}_{1,i}),  
\end{align}
where $\widetilde{\bm{z}}_{1,i} \sim \mathcal{N}(0, \bm{I})$ and it can obtain the deterministic output result $\widetilde{\bm{z}}_{0,i}$.
Finally, we can utilize the decoder ${\rm Dec}(\cdot)$ to generate human motion as $\widetilde{\bm{E}} = {\rm Dec}(\widetilde{\bm{z}}_{0,i})$ accordingly.

\subsubsection{Stochastic Diverse Output Generation Procedure}
Although we have obtained the deterministic ordinary differential equations (DerODE) between the latent space distribution of human motions and the standard Gaussian distribution, it remains challenging to generate highly diverse motion patterns. This difficulty arises from the deterministic nature of the ODEs, as identical initial conditions result in the same output paths, thereby reducing the diversity of the generated samples.
To provide more diverse while accurate human motions, we tend to involve the stochastic differential equations based on the ordinary differential equations in the stochastic diverse output generation procedure. 

\textbf{Proposition 2.}
\textit{Given the stochastic differential equations $d\bm{z}_t = f(\bm{z}_t,t)dt + g (t)d\bm{w}_t$ with the drift and diffusion terms, the mean $\bm{\mu}(t)$ and covariance $\bm{\Sigma}(t)$ can be formulated as:}
%

\begin{equation}
\begin{aligned}
\begin{cases}
&\frac{d \bm{\mu}(t)}{dt}  = \mathbb{E}[f(\bm{z},t)]\\
&\frac{d \bm{\Sigma}(t)}{dt} = \mathbb{E} \left[ f(\bm{z},t) (\bm{z}(t) - \bm{\mu}(t))^{\top} \right] \\
&\quad\quad\quad+ \mathbb{E} \left[ (\bm{z}(t) - \bm{\mu}(t)) f^{\top}(\bm{z},t) \right] + \mathbb{E} [g^2(t)].
\end{cases}
\end{aligned}
\end{equation}

The proof of the Proposition 2 can be found in Appendix \ref{sec:p2}.
We can observe that the stochastic differential equations can transform the distributions according to the specific settings of drift and diffusion terms, which leads to diverse output results based on Proposition 2.
Therefore, it is intuitive to consider a proper stochastic differential equations with carefully designed $f(\bm{z}_t,t)$ and $g(t)$ respectively in the stochastic diverse output generation procedure.

\textbf{Proposition 3.}
\textit{Given the Diverse Stochastic Differential Equations (DivSDE) as $d\bm{z}_t = \left(  -\frac{1}{1-t}\right) \bm{z}_t dt + \eta \sqrt{\frac{2t}{1-t}} \bm{dw}_t$ with the initial data sample $\bm{x}_0$ and the noise level $\eta$, the probability of data distribution $\bm{x}_t$ is $p(\bm{z}_t) = \mathcal{N}((1-t)\bm{z}_i, \eta^2 t^2\bm{I})$ at the time step $t$ when $p(\bm{z}_0) = \mathcal{N}(\bm{z}_0, \bm{0})$.}

The proof of the Proposition 3 can be found in Appendix \ref{sec:p3}.
It is obvious that the diffusion term $\eta \sqrt{\frac{2t}{1-t}} \bm{dw}_t$ which involves noise can enhance the diversity of the model output and thus DivSDE is different than DerODE. 
Note that the stochastic differential equations in Proposition 2 have the backward process as $\bm{dx}_t = \left[  -\frac{1}{1-t}\bm{z}_t - \frac{2t}{1-t} \nabla \log p(\bm{z}_t)\right]  dt + \eta \sqrt{\frac{2t}{1-t} } \bm{dw}_t$, where $\nabla \log p(\bm{z}_t)$ denotes the score function of the data probability.
Specifically, $\nabla \log p(\bm{z}_t)$ can be calculated via $\nabla \log p(\bm{z}_t) = \frac{(1-t)\bm{z}_i - \bm{z}_t}{t^2}$.
Previous score-based approaches \cite{song_score-based_2021} may involve a new neural network to estimate $\nabla \log p(\bm{z}_t)$ even if it is rather time-consuming and hard to train in real practice.
However, it is important to note that we can already obtain $\widetilde{\bm{z}}_{0,i}$ by utilizing DerODE via $\widetilde{\bm{z}}_{0,i} = {\rm DerODE}(\widetilde{\bm{z}}_{1,i})$ and it can be further utilized for DivSDE.
Therefore, we can rewrite the discrete form of the backward process on DivSDE as follows: 
\begin{equation}
\begin{aligned}
\bm{z}_{i,t} &= \bm{z}_{t + \Delta t, i} + \frac{\Delta t}{1-t}\bm{z}_{t + \Delta t, i} \\ &+ \frac{2t\Delta t}{1-t} \frac{(1-t)\widetilde{\bm{z}}_{0,i} - \bm{z}_{t,i}}{t^2} + \eta \varepsilon \sqrt{\frac{2t}{1-t}} \sqrt{\Delta t},
\end{aligned}
\end{equation}
%
where $\varepsilon \sim \mathcal{N}(0, \bm{I})$ denotes the randomly sample noise.
Meanwhile, $\eta$ denotes the strengths of diversity. 
That is, larger value of $\eta$ will provide more diverse output human motions.
Moreover, DivSDE can directly borrow the previously calculated results from DerODE for secondary computations without the need for re-introducing other training processes.

\subsection{Model Summary}  

In summary, the proposed DSDFM can synthesize diverse and accurate human motion sequences through the designed two stages, i.e., \textit{human motion reconstruction} and \textit{diverse motion generation}. 
In the human motion reconstruction stage, we first adopt the human motion reconstruction network to learn a well-structured latent space of human motions through VQVAE network. 
In the diverse motion generation stage, we tend to build the connections between the Gaussian distribution and latent space of human motions, thereby enhancing the diversity while guaranteeing the accuracy of the generated human motions through the designed deterministic feature mapping procedure with DerODE and stochastic diverse output generation procedure with DivSDE. 
Specifically, DerODE can provide deterministic output results in an efficient way.
Meanwhile, DivSDE can obtain more diverse human motions without introducing additional training process.
The pseudo algorithm of the DSDFM is provided in Algorithm \ref{alg1}.

\begin{algorithm} [tb]
  \caption{The process for generating diverse human motions.}
  \label{alg1}
  \begin{algorithmic}[1]
      \REQUIRE
            time interval: $T$, time steps: $\Delta t = \frac{1}{T}$, noise for diversity: $\eta$
      \ENSURE 
        Generated new samples $\hat{E} $.
      \STATE Initialize $\widetilde{\bm{z}}_{1,i}$ from Gaussian distribution $\mathcal{N}(0, \bm{I})$.

       \STATE{ \textcolor{gray}{\# Adopting DerODE to obtain $\widetilde{\bm{x}}_{i,0}$. } }
       
       \STATE{ $\widetilde{\bm{z}}_{0,i} = {\rm DerODE}(\widetilde{\bm{z}}_{1,i})$  }

      \STATE{ \textcolor{gray}{\# Adopting DivSDE to obtain diverse human motions.} }

      \FOR{ $t \in range (T - \Delta t, 0) $}

\STATE Obtain the score function as: $ \nabla \log p(\bm{z}_{t,i}) = \frac{(1-t)\widetilde{\bm{z}}_{0,i} - \bm{z}_{t,i}}{(t/T)^2}.$

        \STATE Obtain the diffusion term as: $ D_{\rm Diffu} = \frac{2(t/T)}{1-(t/T)}  \nabla \log p(\bm{z}_{t,i})  \Delta t $
        \STATE Obtain the drift term as: $ D_{\rm Drift} = \frac{\Delta t}{1-(t/T)}{z}_{ t + \Delta t,i} $
        \STATE Obtain the noise term as: $ 
        \epsilon_{\rm noise} = \eta \cdot \varepsilon \sqrt{\frac{2(t/T)}{1-(t/T)}} \sqrt{\Delta t} $
        \STATE Obtain $ \bm{z}_{t,i} = \bm{z}_{ t+\Delta t,i} + D_{\rm Diffu} + D_{\rm Drift} + \epsilon_{\rm noise}$
      \ENDFOR
      \STATE Generate the human motion: $\hat{E} = {\rm Dec}(\bm{z}_{0,i})   $
  \end{algorithmic}   
\end{algorithm}

%% file: sec/5Experiment.tex
\section{Experiment}
In this section, we provide extensive experiments to evaluate the performance of our proposed DSDFM across widely used human motion datasets. 
We first describe the utilized human motion datasets and implementation details (Section. 5.1). Subsequently, we present a comparative results analysis of our method with other state-of-the-art approaches on conditional and unconditional human motion synthesis. Additionally, we provide ablation studies to assess the effectiveness of the modules in our method (Section 5.2). Finally, we visually showcase the generated diverse human motion sequences to provide a qualitative performance (Section 5.3).

\subsection{Datasets and Implementation Details}
\textbf{Datasets.} The experiments are conducted on two widely used motion capture datasets, i.e., HumanAct12 \cite{guo2020action2motion}, and HumanML3D \cite{guo2022generating}. 
\textbf{HumanAct12} provides 1,191 raw motion sequences, and contains $12$ subjects in which $12$ categories of actions with per-sequence annotation are provided. 
\textbf{HumanML3D} dataset is a recent dataset that contains 14,616 motion sequences annotated by 44970 textual descriptions obtained from AMASS \cite{mahmood2019amass}.

\textbf{Evaluation metrics.}
For a fair comparison, our method employs the following evaluation metrics: Frechet Inception Distance (FID), Kernel Inception Distance (KID), Precision, Recall, Accuracy, Diversity, and Multimodality.
FID is the distance between the feature distribution of generated motions and that of the real motions, namely the difference in mean and variance. 
KID compares skewness as well as the values compared in FID, namely mean and variance. 
Precision measures the probability that a randomly generated motion falls within the support of the distribution of real data. Recall measures the probability that a real motion falls within the support of the distribution of generated data. 
Accuracy is measured by the corresponding action recognition model.
Diversity measures the variance of the whole motion sequences across the dataset. 
Multimodality measures the diversity of human motion generated from the same text description. 
A lower value implies better for FID and KID. Higher Precision, Recall, Accuracy, Diversity, and MultiModality values imply better results.
FID, KID, Precision, Recall, and Accuracy are utilized to evaluate the generated human motion accuracy. Diversity and MultiModality are utilized for the generation diversity.

\textbf{Implementation Details.} 
For the human motion reconstruction process, the VQVAE consists of 4 Transformer layers with 8 heads, and the codebook size is set to 512 $\times$ 512. 
The batch size is set to 128, learning rate is initially set to $10^{-2}$ with a 0.98 decay every 10 epochs. The proposed method is trained for 500 epochs. 
For the diverse motion generation process, the time interval $\Delta t$ is set to 0.01, and the strength of diversity $\eta$ is set to 0.1. The diffusion step is set to 100. 
The balanced parameter $\lambda_{cl}$ for $J_{CL}$ loss is set to 0.3.

\begin{table}[tb]
\centering
\caption{The comparison results of unconditional human motion synthesis between our method and state-of-the-art methods on HumanAct12 dataset. \textbf{Bold} and \underline{underline} indicate the best and the second best result. }
\label{baselineUncon}
\resizebox{\linewidth}{!}{
\begin{tabular}{ccccccc}
\hline
Method & FID $\downarrow$  & KID $\downarrow$ & Precision $\uparrow $  & Recall $\uparrow$    & Diversity $\uparrow$  & \#params \\ \hline
VPoser (CVPR'19)& 48.65 & 0.72 & 0.68      & 0.72   & 12.75   & 29M  \\ 
Action2Motion (MM'21)& 49.76 & 0.68 & 0.70      & 0.71   & 13.80  & 21M   \\ 
ACTOR (CVPR'21)  & 48.80 & 0.53 & \underline{0.72}      & 0.74   & 14.10 & \underline{20M}    \\ 
MDM (ICLR'23)   & 31.92 & 0.96 & 0.66      & 0.62   & 17.00  & 24M  \\ 
MLD (CVPR'23)   & 14.25 & 0.55  & 0.70  & 0.79  & 16.85  & 27M \\ 
Modi (CVPR'23)  & \underline{13.03} & \underline{0.12} & 0.71      & \underline{0.81}   & \underline{17.57}  & 23M  \\ \hline
\textbf{DSDFM (Ours)}   & \textbf{12.86} & \textbf{0.10} & \textbf{0.75}      & \textbf{0.85}   & \textbf{18.41} &  \textbf{15M} \\  \hline
Improvement  & 1.31\%    & 1.67\%   & 4.17\%   & 4.93\%    & 4.78\%  & 2.50\%
\\ \hline
\end{tabular} }
\end{table}

\subsection{Experimental Results}
\textbf{Comparisons on Unconditional Human Motion Synthesis.}
We compare our method DSDFM with other state-of-the-art methods under the unconditional generation settings on the HumanAct12 dataset, the results are shown in Table \ref{baselineUncon}. 
The input of the baseline methods is modified to the same length as our method. 
From the comparison results, we can observe that the baseline methods report poor performance in terms of accuracy and diversity metrics due to the mode collapse or unstable training processes. DSDFM outperforms these methods owing to the designed diverse motion generation procedure. 
In addition, to assess the training efficiency of our method, we also investigate the number of training parameters. The comparison results show that our method utilizes fewer parameters than baseline methods while achieving superior performance, which demonstrates the effectiveness of the proposed method.
This suggests that our method is more computationally efficient and achieves the balance between the accuracy and diversity of generated samples.

\begin{figure*}[htb]
    \centering
    \includegraphics[width=0.9\linewidth]{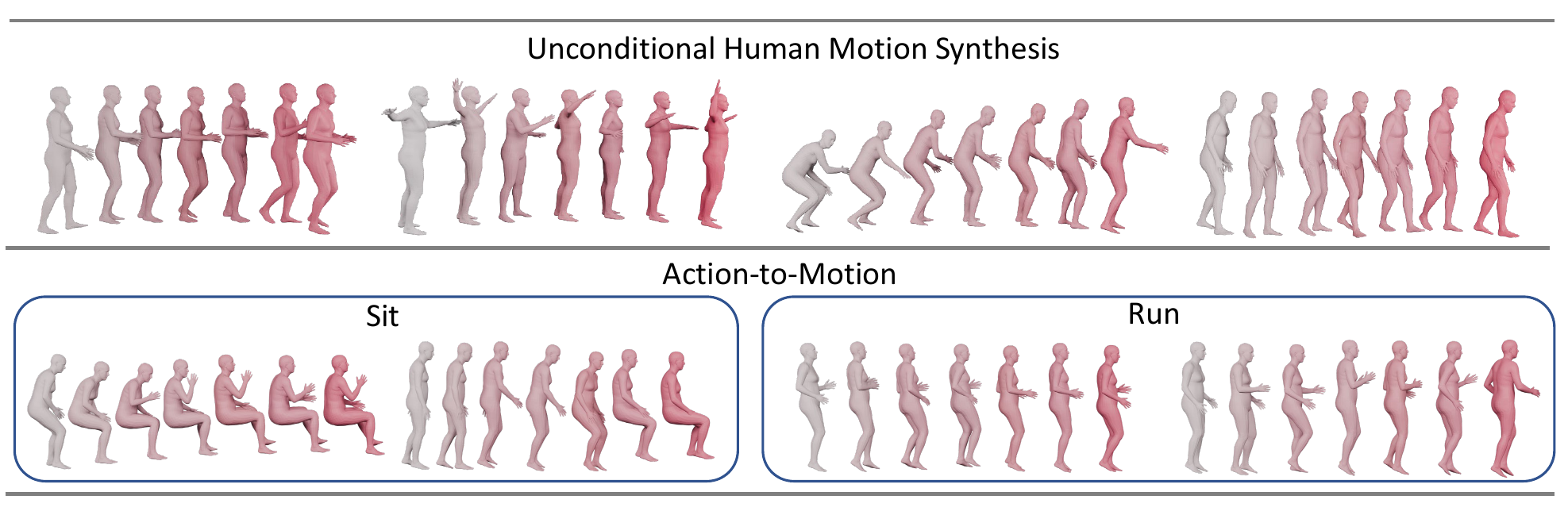}
    \caption{Qualitative results of DSDFM. We present the generated human motion sequences under different settings. The unconditional human motion sequences (top) are generated from the HumanAct12 dataset. The Action-to-Motion results (bottom) show the generated diverse motion sequences under the Sit and Run action labels, which are sampled from the HumanML3D dataset. }
    \label{fig:vis}
\end{figure*}

\begin{figure*}[htpb]
    \centering
    \includegraphics[width=0.68\linewidth]{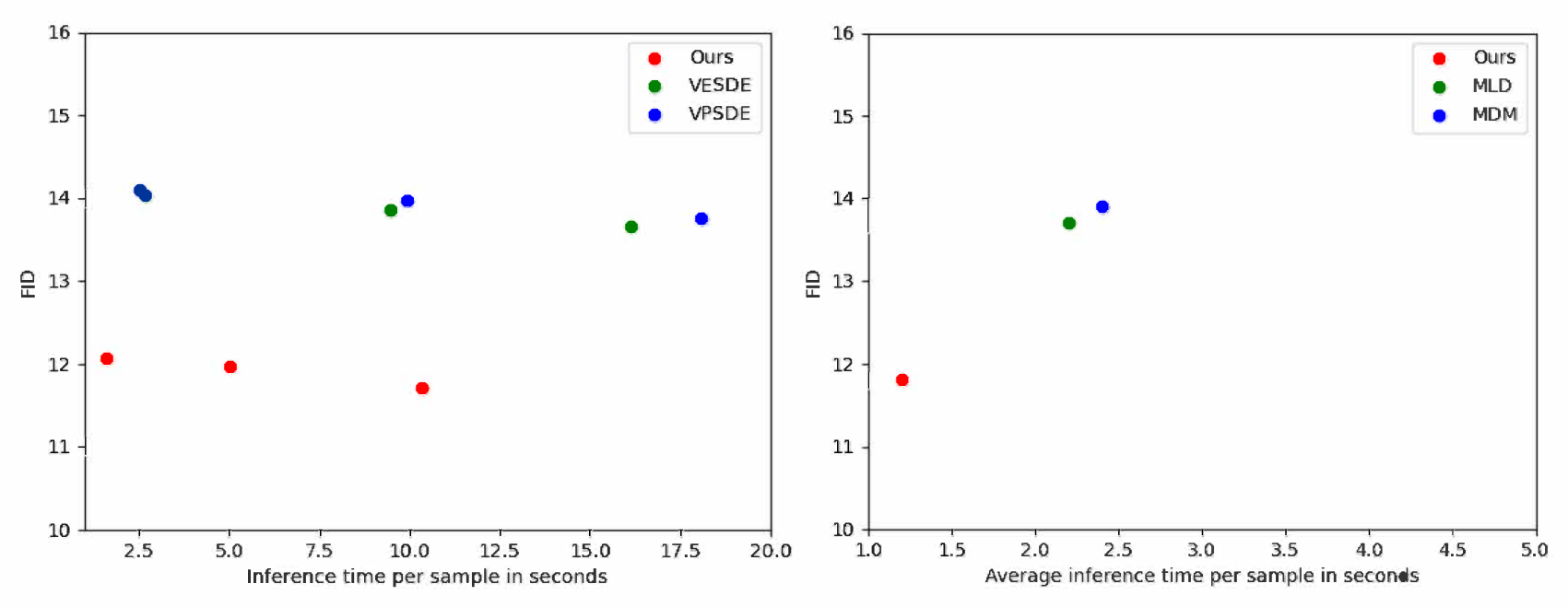}
    \caption{Comparison of the inference time costs of our method on the HumanAct12 dataset. We calculate the ablation studies (left) and average inference time comparison with baselines (right).
    All tests are performed on the NVIDIA A100.
    }
    \label{fig:infer}
\end{figure*}

\textbf{Comparisons on Conditional Human Motion Synthesis}.
Our method can also be extended to conditional generation, i.e., Action-to-Motion task. This task involves generating relevant human motion sequences given an input action label. The comparison results on the HumanAct12 dataset are presented in Table \ref{action2motion}. 
From the comparison results, we can observe that DSDFM also achieves comparable performance under the accuracy and diversity metrics. 
Our method performs slightly worse than MotionDiffuse method in terms of the diversity metric by 0.01\%, but our method reduces the confidence interval, demonstrating that our method is more stable and reliable. Additionally, it significantly decreases the number of training parameters. These results further report the effectiveness of our method for conditional human motion generation.

\begin{table}[]
\centering
\caption{The comparison results of Action-to-Motion task on HumanAct12 dataset. $\pm$ indicates 95\% confidence interval, $\rightarrow$ indicates that closer to real is better. The best results are in bold.  }
\label{action2motion}
\resizebox{1.0\columnwidth}{!}{
\begin{tabular}{cccccc}
\hline
Method   & FID $\downarrow$ & Accuracy $\uparrow$ & Diversity $\rightarrow$ & Multimodality $\uparrow$ & \#params \\ \hline
Real &  $ 0.020^{\pm.010} $ &  $0.997^{\pm.001}$      &  $6.850^{\pm.050} $   & $2.450^{\pm.040} $    &  -     \\ \hline
Action2Motion (MM'21)&      $ 0.338^{\pm.015} $ &    $0.917^{\pm.003}$      &      $6.879^{\pm.066} $    &      $2.511^{\pm.023} $  &  21M     \\ 
ACTOR (CVPR'21) &  $ 0.120^{\pm.000} $ &    $0.955^{\pm.008}$      &      $6.840^{\pm.030} $    &      $2.530^{\pm.020} $ & \underline{20M}   \\ 
INR (ECCV'22) &  $ 0.088^{\pm.004} $ &    $0.973^{\pm.001}$      &      $6.881^{\pm.048} $    &      $2.569^{\pm.040} $   &  25M   \\ 
MLD (CVPR'23) & $ 0.077^{\pm.004} $ &    $0.964^{\pm.002}$      &      $6.831^{\pm.050} $    &      $2.824^{\pm.0.38} $   &  27M   \\ 
MDM (ICLR'23) &  $ 0.100^{\pm.000} $ &    $0.990^{\pm.000}$      &      $6.860^{\pm.050} $    &      $2.520^{\pm.010} $  &  24M      \\
MotionDiffuse (TPAMI'24) &  $ \underline{0.070}^{\pm.000} $ &    $\underline{0.992}^{\pm.013}$      &      $\textbf{6.850}^{\pm.020} $    &      $\underline{2.460}^{\pm.020} $  &  25M      \\
\hline
\textbf{DSDFM (Ours)}   & $\textbf{0.068}^{\pm.010}$  &    $\textbf{0.994}^{\pm.001}$      &      $\underline{6.851}^{\pm.008} $    &      $\textbf{2.455}^{\pm.025} $  & \textbf{15M } \\ \hline
Improvement  & 2.85\%    & 0.21 \%  & -0.01\%  &  0.21\%   &  2.50\%  
\\ \hline
\end{tabular}   }  
\end{table}

\begin{table}[]
\centering
\caption{Ablation study on the comparison results of training and inference time on the HumanAct12 dataset. $m$ denotes minute, $s$ denotes second. }
\label{trainingtime1}
\resizebox{\linewidth}{!}{
\begin{tabular}{cccccc}
\hline
\multirow{2}{*}{Method} & \multirow{2}{*}{Epoch} & \multirow{2}{*}{Training Time (m)} & \multicolumn{3}{c}{Inference Time (s)/FID}     \\ \cline{4-6} 
                        &                        &                                    & 100 steps/FID & 500 steps/FID & 1000 steps/FID \\ \hline
VPSDE                   & 500                    & 42.93                              & 2.54/16.74    & 9.93/15.63    & 18.09/14.31    \\
VESDE                   & 500                    & 40.57                              & 2.68/16.49    & 9.48/14.92    & 16.12/14.17    \\ \hline
DSDFM(Ours)             & 500                    & 25.33                              & 1.60/13.61    & 5.03/12.86    & 10.33/12.24    \\ \hline
\end{tabular}  }
\end{table}

\subsection{Ablation studies}
To report the effectiveness of each component of our method, we compare the baseline methods with DSDFM under different settings on the HumanML3D and HumanAct12 datasets, including the training time, inference time for different diffusion steps, and the corresponding FID. The results are shown in Table \ref{trainingtime1} and Table \ref{trainingtime2}.
Table \ref{trainingtime1} shows the comparison results for unconditional motion generation on the HumanAct12 dataset. 
From the results, we can observe that, compared to VPSDE and VESDE given the same number of epochs, DSDFM significantly reduces the training time while achieving a comparable performance under the FID metric, which demonstrates that our method is easier to train than the baseline methods and guaranteeing the quality of generated human motions. 
We also test these methods under different diffusion steps, and the performance of our method is improved in inference time. 
In addition, Table \ref{trainingtime2} shows the comparison results for Action-to-Motion task on HumanML3D dataset. The results under the same metrics are consistent with the results on the HumanAct12 dataset, which further demonstrates the effectiveness of our method.

\begin{table}[]
\centering
\caption{Ablation study on the comparison results of training and inference time on the HumanML3D dataset.}
\label{trainingtime2}
\resizebox{\linewidth}{!}{
\begin{tabular}{cccccc}
\hline
\multirow{2}{*}{Method} & \multirow{2}{*}{Epoch} & \multirow{2}{*}{Training Time (m)} & \multicolumn{3}{c}{Inference Time (s)/FID}                                              \\ \cline{4-6} 
                        &                        &                                    & 100 steps/FID & 500 steps/FID                                          & 1000 steps/FID \\ \hline
VPSDE                   & 500                    & 12.54                              & 2.47/ $0.092^{\pm.003}$        & 4.95/                $0.088^{\pm.002}$                                 & 6.51/ $0.080^{\pm.024}$          \\
VESDE                   & 500                    & 12.57                              & 2.35/ $0.094^{\pm.005}$         & 4.48/ $0.089^{\pm.001}$                                                 & 6.62/ $0.078^{\pm.013}$          \\ \hline
DSDFM(Ours)             & 500                    & 7.02                               & 1.01/$0.073^{\pm.005}$         & 2.15/$0.068^{\pm.008}$  & 4.82/$0.054^{\pm.010}$          \\ \hline
\end{tabular}  }
\end{table}

\begin{table}[]
\centering
\caption{Ablation studies of the proposed method. We compare our method with other score-based methods and provide the comparison results under the accuracy and diversity metrics, as well as the number of training parameters. }
\label{ablation}
\resizebox{\linewidth}{!}{
\begin{tabular}{ccccccc}
\hline
Method & FID $\downarrow$  & KID $\downarrow$ & Precision $\uparrow $  & Recall $\uparrow$    & Diversity $\uparrow$ & \#param \\ \hline
VESDE      &  14.92   &  0.36       & 0.59     & 0.65   & 16.21    & 28M  \\
VPSDE      & 15.63     & 0.29       & 0.64     & 0.68   & 17.00  & 24M  \\
SDE (DDPM++)  & 13.25    &  0.21    & 0.68     & 0.75   & 17.46    & 22M    \\
SDE (NCSN++)  & 13.01    &   0.19   & 0.72     & 0.79   & 17.54     & 21M    \\ \hline
\textbf{DSDFM (Ours)}   & \textbf{12.86} & \textbf{0.10} & \textbf{0.75}      & \textbf{0.85}   & \textbf{18.41}   & \textbf{15M } \\ \hline
\end{tabular} }
\end{table}

The ablation studies also test the performance of the designed stochastic diverse output generation procedure in DSDFM under the diversity and accuracy metrics, the results are shown in Table \ref{ablation}. Specifically, we employ other score-based methods to enhance the diversity of generated human motion sequences, i.e., variance preserving SDE (VPSDE), variance exploding SDE (VESDE), DDPM++, and NCSN++. From the comparison results, we can observe that our method exhibits comparable performance in terms of accuracy compared to the baseline methods, while showing a slight improvement in diversity. Notably, we have achieved a significant reduction in the number of training parameters, which report the effectiveness of our method.

\subsection{Visualization}
In this section, we show the visualization results of our method on the unconditional human motion synthesis and Action-to-Motion tasks. As depicted in Figure \ref{fig:vis}, the top is the unconditional human motion synthesis, all human motion sequences are unconditionally generated from random noise sampled from Gaussian distribution on the HumanAct12 dataset. The figure shows that our method can generate diverse and high fidelity human motion sequences. 
The bottom is the sequences generated by the Action-to-Motion task, the generated sequences are under the action labels on the HumanML3D dataset. We can observe that the generated diverse motion sequences match the descriptions well.
These qualitative results demonstrate that DSDFM can generate diverse and coherent human motion sequences.

We compare and visualize the comparison results of inference time in Figure \ref{fig:infer}. The left of this Figure is the ablation studies of our method with different diffusion steps. This figure shows that using VPSDE and VESDE as our backbone has long inference time and relatively low accuracy. 
The right of this Figure is the average inference time comparison with baselines, which shows that our method can speed up the inference time when generating new samples.

%% file: sec/6Conclusion.tex
\section{Conclusion}
In this paper, we propose a Deterministic-to-Stochastic Diverse Latent Feature Mapping (DSDFM) for human motion synthesis. DSDFM is easy to train compared with the recent SGMs-based method, while facilitating the diversity and accuracy of generated human motions.
DSDFM includes two stages, human motion reconstruction and diverse motion generation. 
Human motion reconstruction aims to learn a well-structured latent space of human motions.
Diverse motion generation aims to enhance the diversity of the generated human motion sequences through the designed deterministic feature mapping procedure with DerODE and stochastic diverse output generation procedure with DivSDE.
Extensive experimental results demonstrate the efficacy of the proposed DSDFM method for human motion synthesis.

%% file: sec/appendix.tex
\clearpage


\appendix

\section*{\centerline{Appendix}}
\section{Calculation on Optimal Transport}
\label{sec:p1}

In this section, we will provide the optimize details on optimal transport.
That is, the problem definition of optimal transport is given as:
\begin{equation} 
\begin{aligned}
\left.
\begin{aligned}
&\min_{\boldsymbol{\pi} \in \Delta} J_{\rm OT} = \langle \boldsymbol{\pi}, \boldsymbol{C} \rangle  \\&s.t.\,\, \Delta = \left\{ \sum_{j=1}^N \pi_{ij} = a_i,\quad \sum_{i=1}^N \pi_{ij} = b_j, \quad \pi_{ij} \ge 0\right\},
\end{aligned}
\right.  
\end{aligned}
\end{equation} 
To start with, we should first figure out the Lagrange multipliers of optimal transport as:
\begin{equation}
\small
\begin{aligned}
\max_{\boldsymbol{f}, \boldsymbol{g}, \boldsymbol{s}} \min_{\boldsymbol{\pi}} \mathcal{J}  =   \langle \boldsymbol{f},  \boldsymbol{a}\rangle + \langle \boldsymbol{g},  \boldsymbol{b}\rangle +   \left[ \sum_{i,j}  \left(   C_{ij} - f_i - g_j   - s_{ij} \right)\pi_{ij}  \right]  
\end{aligned}
\end{equation}
where $\bm{f}$, $\bm{g}$ and $\bm{s}$ denote the Lagrange multipliers.
By taking the differentiation on $\pi_{ij}$, we can obtain the following results as:
\begin{equation}
\begin{aligned}
\left\{
\begin{aligned}
& \frac{\partial \mathcal{J}}{\partial \pi_{ij}}  = C_{ij} - f_i - g_j   - s_{ij} = 0\\
& s_{ij}  \ge 0
\end{aligned}
\right. 
\end{aligned}
\end{equation}
Note that $s_{ij} \ge 0$ and $s_{ij}\pi_{ij} = 0$ according to the KKT condition.
Therefore, we obtain the dual form of optimal transport:
\begin{equation}
\begin{aligned}
&\max_{\boldsymbol{f}, \boldsymbol{g}}  \mathcal{J}_{\rm OT}  = \langle \boldsymbol{f},  \boldsymbol{a}\rangle + \langle \boldsymbol{g},  \boldsymbol{b}\rangle  \\
& s.t.\,\, f_i + g_j \le C_{ij}
\end{aligned}
\end{equation} 
Specifically, we can adopt the \textbf{$c$-transform} via $g_j = \inf_{k\in [M]} \left( C_{kj} - f_k  \right)$.
Meanwhile the optimal transport can be transformed into the following convex optimization problem:
\begin{equation}
\small
\begin{aligned}
\label{equ:jot}
\mathcal{J}_{\rm OT} = \arg \max_{\boldsymbol{f} } \left[ \sum_{i=1}^N f_i   a_i + \sum_{j=1}^N \left[ \inf_{k\in [N]} \left( C_{kj} - f_k  \right) \right]   b_j \right]
\end{aligned}
\end{equation} 
We can adopt commonly-used optimization methods (e.g., L-BFGS) to obtain the optimal solution on $\bm{f}$.
After we obtain the optimal result on $\bm{f}^*$, we can obtain $\bm{s}$ accordingly:
\begin{equation} 
\begin{aligned}
\label{equ:jot}
s_{ij} = C_{ij} - f^*_i - \inf_{k\in [N]} \left( C_{kj} - f^*_k  \right)
\end{aligned}
\end{equation} 
Since we set $a_i = b_j = \frac{1}{N}$, the matching results in $\pi_{ij}$ can be obtained as:
\begin{equation} 
\begin{aligned}
\label{equ:jot}
\pi_{ij} = \left\{
\begin{aligned}
& \frac{1}{N}, \quad s_{ij} = 0 \\
& 0, \quad \quad  s_{ij} > 0
\end{aligned}
\right.
\end{aligned}
\end{equation}

\section{Proof of Proposition 2}
\label{sec:p2}

\textbf{Proposition 2.}
\textit{Given the stochastic differential equations $d\bm{z}_t = f(\bm{z}_t,t)dt + g (t)d\bm{w}_t$ with the drift and diffusion terms, the mean $\bm{\mu}(t)$ and covariance $\bm{\Sigma}(t)$ can be formulated as:}
\begin{equation}
\begin{aligned}
\begin{cases}
&\frac{d \bm{\mu}(t)}{dt}  = \mathbb{E}[f(\bm{z},t)]\\
&\frac{d \bm{\Sigma}(t)}{dt} = \mathbb{E} \left[ f(\bm{z},t) (\bm{z}(t) - \bm{\mu}(t))^{\top} \right] \\
&\quad\quad\quad+ \mathbb{E} \left[ (\bm{z}(t) - \bm{\mu}(t)) f^{\top}(\bm{z},t) \right] + \mathbb{E} [g^2(t)]
\end{cases}
\end{aligned}
\end{equation}

\begin{proof}

To start with, it is noticeable that the mean value of the diffusion term $d\bm{w}_t$ is 0. Therefore, it is easy to verify that $\frac{d \bm{\mu}(t)}{dt}  = \mathbb{E}[f(\bm{z},t)]$.
Meanwhile, the covariance term can be figure out as:
\begin{equation} 
\label{equ:64}
\begin{aligned} 
&d\bm{\Sigma}(t) = \mathbb{E}[d[(\bm{z}(t) - \bm{\mu}(t))(\bm{z}(t) - \bm{\mu}(t))^\top]] \\&= \mathbb{E}[d(\bm{z} - \bm{\mu})(\bm{z} - \bm{\mu})^\top + (\bm{z} - \bm{\mu})d(\bm{z} - \bm{\mu})^\top + d(\bm{z} - \bm{\mu})d(\bm{z} - \bm{\mu})^\top]
\end{aligned} 
\end{equation}
%
To simplify the first term, we should notice that:
\begin{equation}  
\label{equ:62}
\begin{aligned}
&\mathbb{E} \left[ (d\bm{z}(t) - d\bm{\mu}(t)) (\bm{z}(t) - \bm{\mu}(t))^{\top}\right] \\& = \mathbb{E} \left[ (d\bm{z}(t) -  \mathbb{E} \left[ f(\bm{z}, t)\right]  dt ) (\bm{z}(t) - \bm{\mu}(t))^{\top}\right] \\
& = \mathbb{E} \left[ d\bm{z}(t)  (\bm{z}(t) - \bm{\mu}(t))^{\top}\right]
\end{aligned} 
\end{equation}
To simplify the second term, we also have the results as:
\begin{equation} 
\label{equ:64}
\begin{aligned} 
\mathbb{E}[d(\bm{z} - \bm{\mu})d(\bm{z} - \bm{\mu})^\top] &= \mathbb{E}\left[(g(t)d\bm{w}_t)(g(t)d\bm{w}_t)^\top \right] \\&= \mathbb{E}[g^2(t)]dt
\end{aligned} 
\end{equation}

%
Therefore, we have obtain the final solution: 
\begin{equation}  
\label{equ:62}
\begin{aligned}
d\bm{\Sigma}(t) &=  \mathbb{E} \left[ (d\bm{z}(t) - d\bm{\mu}(t)) (\bm{z}(t) - \bm{\mu}(t))^{\top}\right] \\ & + \mathbb{E} \left[ ( \bm{z}(t) -  \bm{\mu}(t)) (d\bm{z}(t) - d\bm{\mu}(t))^{\top}\right] +    \mathbb{E} [g^2(t)] dt    \\
& = \mathbb{E} \left[ f(\bm{z},t) (\bm{z}(t) - \bm{\mu}(t))^{\top}\right]dt \\&+ \mathbb{E} \left[ ( \bm{z}(t) -  \bm{\mu}(t))\left(f(\bm{z},t) \right)^{\top} \right]dt +    \mathbb{E} [g^2(t)] dt
\end{aligned} 
\end{equation}
\end{proof}

\section{Proof of Proposition 3}
\label{sec:p3}

\textbf{Proposition 3.}
\textit{Given the Diverse Stochastic Differential Equations (DivSDE) as $d\bm{x}_t = \left(  -\frac{1}{1-t}\right) \bm{x}_t dt + \eta \sqrt{\frac{2t}{1-t}} \bm{dw}_t$ with the initial data sample $\bm{z}_0$ and the noise level $\eta$, the probability of data distribution $\bm{z}_t$ is $p(\bm{x}_t) = \mathcal{N}((1-t)\bm{z}_t, \eta^2 t^2\bm{I})$ at the time step $t$ when $p(\bm{z}_0) = \mathcal{N}(\bm{z}_0, \bm{0})$.}

\begin{proof}
Adopting the Proposition 2, we can provide the equations on mean and covariance as below:
\begin{equation} 
\left\{
\begin{aligned}
& \frac{d \bm{\mu}(t)}{dt} = \left(  -\frac{1}{1-t}\right) \bm{\mu}(t) \\
& \frac{d \bm{\Sigma} (t)}{dt} = \left(  -\frac{2}{1-t}\right)\bm{\Sigma} (t)  + \eta^2 \frac{2t}{1-t} 
\end{aligned}
\right.
\end{equation}

The solutions are given as $\bm{\mu}(t) = (1-t)\bm{z}_i$ and $\bm{\Sigma}(t) = \eta^2 t^2 \bm{I}$. 
\end{proof}

\begin{figure*}
    \centering
    \includegraphics[width=1\linewidth]{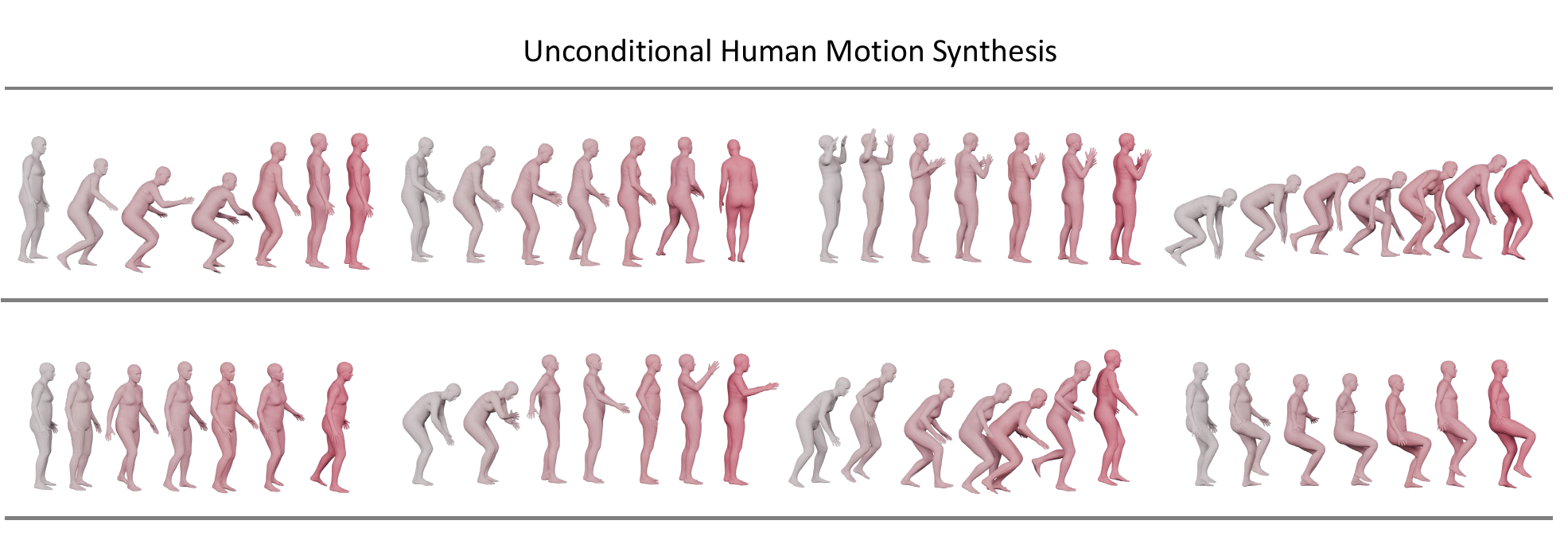}
    \caption{Qualitative results of DSDFM. We present more generated unconditional human motion sequences. }
    \label{fig:vis1}
\end{figure*}

\begin{figure*}
    \centering
    \includegraphics[width=1\linewidth]{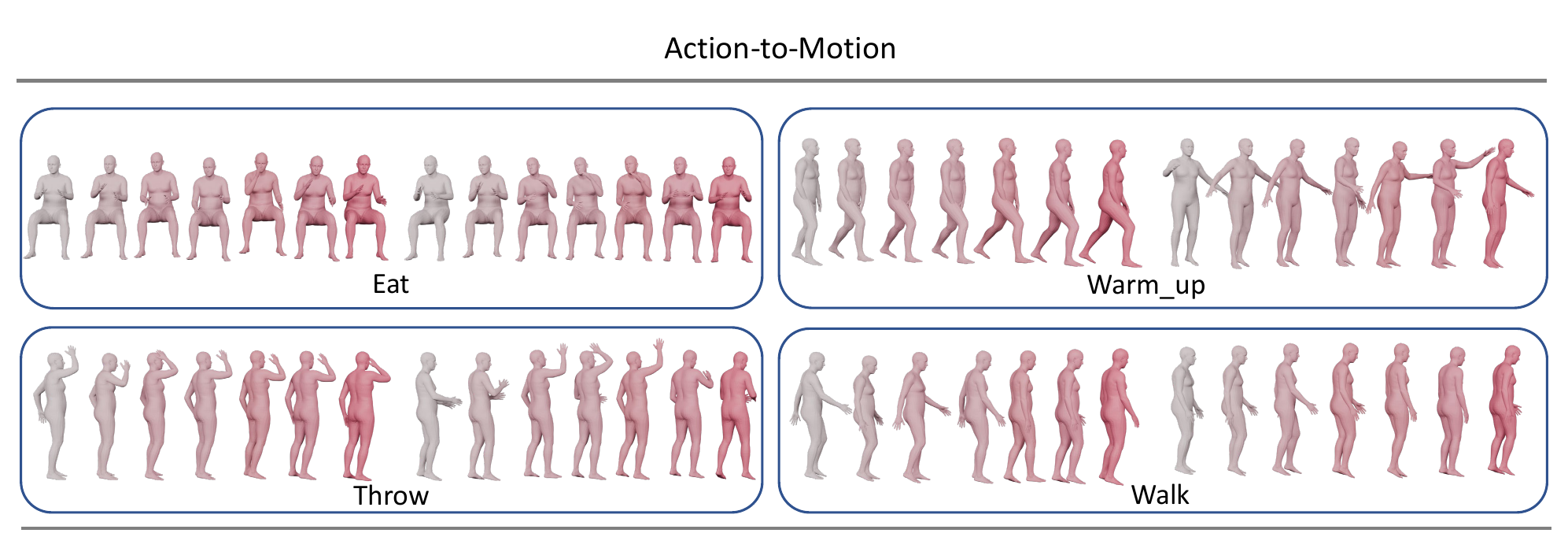}
    \caption{Qualitative results of DSDFM. We present the diverse human motion sequences under different actions. }
    \label{fig:vis2}
\end{figure*}

\begin{figure*}
    \centering
    \includegraphics[width=0.90\linewidth]{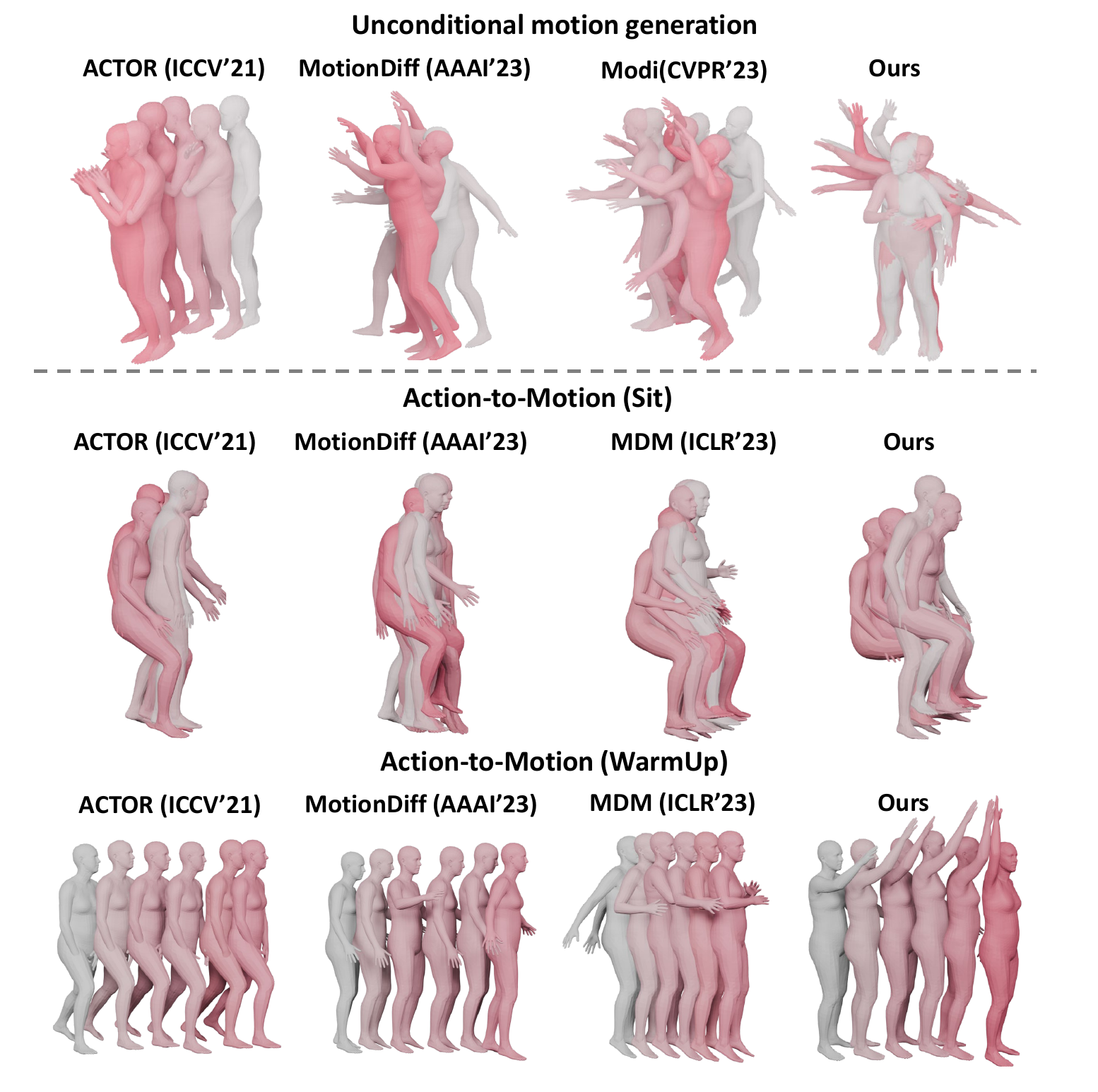} 
    \caption{The qualitative comparison results of the state-of-the-art methods and our proposed DSDFM. }
    \label{fig:ComVis}
\end{figure*}

\section{Experiment Results }
\subsection{Metric Definitions}
In this work, we use the following metrics to measure the performance of the proposed method for unconditional human motion synthesis and Action-to-Motion tasks.

\textbf{Frechet Inception Distance (FID)}. FID calculates the distribution distance between the generated and real motions. FID is an important metric widely used to evaluate the overall quality of generated motions. The FID is calculated as:
\begin{equation}
    \mathrm{FID}=\|\mu_{gt}-\mu_{pred}\|^2-\mathrm{Tr}(\Sigma_{gt}+\Sigma_{pred}-2(\Sigma_{gt}\Sigma_{pred})^{\frac12}),
\end{equation}
where $\Sigma$ is the covariance matrix. $Tr$ denotes the trace of a matrix. $\mu_{gt}$ and $\mu_{pred}$ are the mean of
ground-truth motion features and generated motion features.

\textbf{Kernel Inception Distance (KID)}. KID compares skewness as well as the values compared in FID \cite{guo2022generating}, namely mean and variance. KID is known to work better for small and medium size datasets. 

\textbf{Precision, Recall}. These measures are adopted from the discriminative domain to the generative domain \cite{sajjadi2018assessing}. Precision measures the probability that a randomly generated motion falls within the support of the distribution of real images, and is closely related with fidelity. Recall measures the probability that a real motion falls within the support of the distribution of generated images, and is closely related with diversity.

\textbf{Accuracy}. We use a pre-trained action recognition classifier \cite{guo2020action2motion} to classify human motions and calculate the overall recognition accuracy. The recognition accuracy indicates the correlation between the motion and its action type.

\textbf{Diversity}. Diversity measures the variance of the generated motions across all action categories. From a set of all generated motions from various action types, two subsets of the same size $S_d$ are randomly sampled. Their respective sets of motion feature vectors $\left\{\mathbf{v}_{1},\cdots,\mathbf{v}_{{S_{d}}}\right\}$ and $\left\{\mathbf{v}_{1}^{\prime},\cdots,\mathbf{v}_{{S_{d}^{\prime}}}\right\}$ are extracted. The diversity of this set of motions is defined as:
\begin{equation}
    \mathrm{Diversity}=\frac{1}{S_d}\sum_{i=1}^{S_d}\parallel\mathrm{v}_i-\mathrm{v}_i'\parallel_2.
\end{equation}
where $S_d=200$ is used in experiments.

\textbf{Multimodality}. Different from diversity, multimodality measures how much the generated motions diversify within each action type. Given a set of motions with $C$ action types. For $c$-th action, we randomly sample two subsets with the same size $S_l$, and then extract two subsets of feature vectors $\{\mathrm{v}_{c,1},\cdots,\mathrm{v}_{c, S_{l}}\}$ and $\{\mathrm{v}_{c,1}^{\prime},\cdots,\mathrm{v}_{c, S_l}^{\prime}\}.$The multimodality of this motion set is formalized as:
\begin{equation}
    \text{Multimodality}=\frac{1}{C\times S_{l}}\sum_{c=1}^{C}\sum_{i=1}^{S_{l}}\left\|\mathrm{v}_{c,i}-\mathrm{v'}_{c,i}\right\|_{2}.
\end{equation}
where $S_l=20$ is used in experiments

\begin{figure}
    \centering
    \includegraphics[width=1\linewidth]{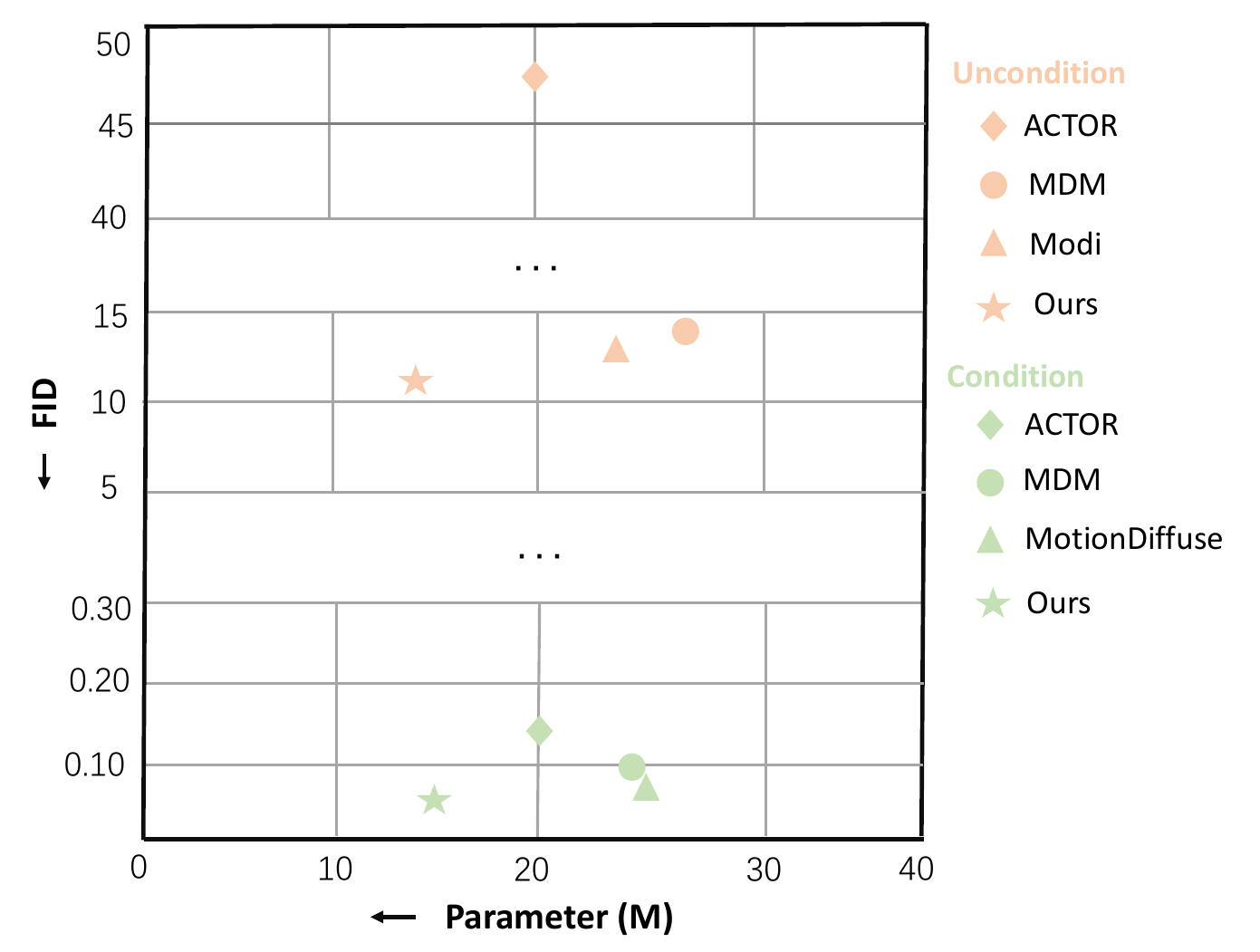}
    \caption{Comparison of the training parameter and the corresponding FID metric. }
    \label{fig:TrainingTimeCom}
\end{figure}

\subsection{Additional Visualization Results}
We provide additional visualization of human motion results in this section, which consists of the unconditional human motion synthesis and Action-to-Motion tasks.

\textbf{Qualitative Analysis on Unconditional Human Motion Synthesis}.
Figure \ref{fig:vis1} visualizes a broader range of unconditional human motion sequences, effectively highlighting the remarkable diversity and high fidelity achieved by our proposed DSDFM.
The visualization results demonstrate the remarkable ability of our method to produce diverse and realistic human motion sequences in unconditional human motion synthesis task.

\textbf{Qualitative Analysis on Action-to-Motion}.
Figure \ref{fig:vis2} illustrates diverse human motion sequences across various action categories, providing evidence that our method is comparable under different action conditions.

\textbf{ Comparison with Other Methods.} 
We provide more qualitative comparison of the state-of-the-art methods on human motion synthesis, i.e, unconditional motion generation and conditional motion generation under action labels (Action-to-Motion). 
As shown in Figure \ref{fig:ComVis}, we compare our method with the state-of-the-art methods. 
Under unconditional generation, the visual results of other methods show that the generated motion sequences tend to converge to static poses, resulting in a lack of diversity. Under action label conditional generation, some methods generate sequences that fail to meet the semantic requirements. The comparison results show that our method can achieve more diverse and accurate human motion sequences. 
More visualization results of our method can be seen in the supplementary video. These extensive results indicate that our method not only enables a significantly faster training process but also produces motion sequences with greater fidelity.

In addition, we visualize the comparison results of the training parameter and the corresponding FID metric. As shown in Figure \ref{fig:TrainingTimeCom}. Our method achieves the best performance while utilizing the fewest training parameters. These results further underscore the effectiveness of the proposed approach.

%% file: main.bbl
\begin{thebibliography}{61}
\providecommand{\natexlab}[1]{#1}
\providecommand{\url}[1]{\texttt{#1}}
\expandafter\ifx\csname urlstyle\endcsname\relax
  \providecommand{\doi}[1]{doi: #1}\else
  \providecommand{\doi}{doi: \begingroup \urlstyle{rm}\Url}\fi

\bibitem[Alexanderson et~al.(2023)Alexanderson, Nagy, Beskow, and Henter]{alexanderson2023listen}
Simon Alexanderson, Rajmund Nagy, Jonas Beskow, and Gustav~Eje Henter.
\newblock Listen, denoise, action! audio-driven motion synthesis with diffusion models.
\newblock \emph{ACM Transactions on Graphics (TOG)}, 42\penalty0 (4):\penalty0 1--20, 2023.

\bibitem[Cervantes et~al.(2022)Cervantes, Sekikawa, Sato, and Shinoda]{cervantes2022implicit}
Pablo Cervantes, Yusuke Sekikawa, Ikuro Sato, and Koichi Shinoda.
\newblock Implicit neural representations for variable length human motion generation.
\newblock In \emph{European Conference on Computer Vision}, pages 356--372. Springer, 2022.

\bibitem[Chen et~al.(2023)Chen, Jiang, Liu, Huang, Fu, Chen, and Yu]{chen2023executing}
Xin Chen, Biao Jiang, Wen Liu, Zilong Huang, Bin Fu, Tao Chen, and Gang Yu.
\newblock Executing your commands via motion diffusion in latent space.
\newblock In \emph{Proceedings of the IEEE/CVF Conference on Computer Vision and Pattern Recognition}, pages 18000--18010, 2023.

\bibitem[Chung et~al.(2014)Chung, Gulcehre, Cho, and Bengio]{chung2014empirical}
Junyoung Chung, Caglar Gulcehre, KyungHyun Cho, and Yoshua Bengio.
\newblock Empirical evaluation of gated recurrent neural networks on sequence modeling.
\newblock \emph{arXiv preprint arXiv:1412.3555}, 2014.

\bibitem[Davtyan et~al.(2023)Davtyan, Sameni, and Favaro]{davtyan2023efficient}
Aram Davtyan, Sepehr Sameni, and Paolo Favaro.
\newblock Efficient video prediction via sparsely conditioned flow matching.
\newblock In \emph{Proceedings of the IEEE/CVF International Conference on Computer Vision}, pages 23263--23274, 2023.

\bibitem[Dhariwal and Nichol(2021)]{dhariwal2021diffusion}
Prafulla Dhariwal and Alexander Nichol.
\newblock Diffusion models beat gans on image synthesis.
\newblock \emph{Advances in neural information processing systems}, 34:\penalty0 8780--8794, 2021.

\bibitem[Diomataris et~al.(2024)Diomataris, Athanasiou, Taheri, Wang, Hilliges, and Black]{diomataris2024wandr}
Markos Diomataris, Nikos Athanasiou, Omid Taheri, Xi Wang, Otmar Hilliges, and Michael~J Black.
\newblock Wandr: Intention-guided human motion generation.
\newblock In \emph{Proceedings of the IEEE/CVF Conference on Computer Vision and Pattern Recognition}, pages 927--936, 2024.

\bibitem[Gao et~al.(2020)Gao, Song, Poole, Wu, and Kingma]{gao2020learning}
Ruiqi Gao, Yang Song, Ben Poole, Ying~Nian Wu, and Diederik~P Kingma.
\newblock Learning energy-based models by diffusion recovery likelihood.
\newblock \emph{arXiv preprint arXiv:2012.08125}, 2020.

\bibitem[Guo et~al.(2020)Guo, Zuo, Wang, Zou, Sun, Deng, Gong, and Cheng]{guo2020action2motion}
Chuan Guo, Xinxin Zuo, Sen Wang, Shihao Zou, Qingyao Sun, Annan Deng, Minglun Gong, and Li Cheng.
\newblock Action2motion: Conditioned generation of 3d human motions.
\newblock In \emph{Proceedings of the 28th ACM International Conference on Multimedia}, pages 2021--2029, 2020.

\bibitem[Guo et~al.(2022)Guo, Zou, Zuo, Wang, Ji, Li, and Cheng]{guo2022generating}
Chuan Guo, Shihao Zou, Xinxin Zuo, Sen Wang, Wei Ji, Xingyu Li, and Li Cheng.
\newblock Generating diverse and natural 3d human motions from text.
\newblock In \emph{Proceedings of the IEEE/CVF Conference on Computer Vision and Pattern Recognition}, pages 5152--5161, 2022.

\bibitem[Guo et~al.(2024)Guo, Mu, Javed, Wang, and Cheng]{guo2024momask}
Chuan Guo, Yuxuan Mu, Muhammad~Gohar Javed, Sen Wang, and Li Cheng.
\newblock Momask: Generative masked modeling of 3d human motions.
\newblock In \emph{Proceedings of the IEEE/CVF Conference on Computer Vision and Pattern Recognition}, pages 1900--1910, 2024.

\bibitem[Ho et~al.(2020)Ho, Jain, and Abbeel]{ho2020denoising}
Jonathan Ho, Ajay Jain, and Pieter Abbeel.
\newblock Denoising diffusion probabilistic models.
\newblock \emph{Advances in Neural Information Processing Systems}, 33:\penalty0 6840--6851, 2020.

\bibitem[Holden et~al.(2016)Holden, Saito, and Komura]{holden2016deep}
Daniel Holden, Jun Saito, and Taku Komura.
\newblock A deep learning framework for character motion synthesis and editing.
\newblock \emph{ACM Transactions on Graphics (TOG)}, 35\penalty0 (4):\penalty0 1--11, 2016.

\bibitem[Jin et~al.(2024)Jin, Wu, Fan, Sun, Yang, and Yuan]{jin2024act}
Peng Jin, Yang Wu, Yanbo Fan, Zhongqian Sun, Wei Yang, and Li Yuan.
\newblock Act as you wish: Fine-grained control of motion diffusion model with hierarchical semantic graphs.
\newblock \emph{Advances in Neural Information Processing Systems}, 36, 2024.

\bibitem[Li et~al.(2021{\natexlab{a}})Li, Kang, Pei, Zhe, Zhang, He, and Bao]{li2021audio2gestures}
Jing Li, Di Kang, Wenjie Pei, Xuefei Zhe, Ying Zhang, Zhenyu He, and Linchao Bao.
\newblock Audio2gestures: Generating diverse gestures from speech audio with conditional variational autoencoders.
\newblock In \emph{Proceedings of the IEEE/CVF International Conference on Computer Vision}, pages 11293--11302, 2021{\natexlab{a}}.

\bibitem[Li et~al.(2021{\natexlab{b}})Li, Yang, Ross, and Kanazawa]{li2021ai}
Ruilong Li, Shan Yang, David~A Ross, and Angjoo Kanazawa.
\newblock Ai choreographer: Music conditioned 3d dance generation with aist++.
\newblock In \emph{Proceedings of the IEEE/CVF International Conference on Computer Vision}, pages 13401--13412, 2021{\natexlab{b}}.

\bibitem[Li et~al.(2021{\natexlab{c}})Li, Yang, Ross, and Kanazawa]{li2021learn}
Ruilong Li, Shan Yang, David~A Ross, and Angjoo Kanazawa.
\newblock Learn to dance with aist++: Music conditioned 3d dance generation.
\newblock \emph{arXiv preprint arXiv:2101.08779}, 2\penalty0 (3), 2021{\natexlab{c}}.

\bibitem[Lin et~al.(2023)Lin, Zeng, Lu, Cai, Zhang, Wang, and Zhang]{NEURIPS2023_4f8e27f6}
Jing Lin, Ailing Zeng, Shunlin Lu, Yuanhao Cai, Ruimao Zhang, Haoqian Wang, and Lei Zhang.
\newblock Motion-x: A large-scale 3d expressive whole-body human motion dataset.
\newblock In \emph{Advances in Neural Information Processing Systems}, pages 25268--25280. Curran Associates, Inc., 2023.

\bibitem[Lipman et~al.(2022)Lipman, Chen, Ben-Hamu, Nickel, and Le]{lipman2022flow}
Yaron Lipman, Ricky~TQ Chen, Heli Ben-Hamu, Maximilian Nickel, and Matt Le.
\newblock Flow matching for generative modeling.
\newblock \emph{arXiv preprint arXiv:2210.02747}, 2022.

\bibitem[Lipman et~al.(2023)Lipman, Chen, Ben-Hamu, Nickel, and Le]{lipman2023flow}
Yaron Lipman, Ricky T.~Q. Chen, Heli Ben-Hamu, Maximilian Nickel, and Matthew Le.
\newblock Flow matching for generative modeling.
\newblock In \emph{The Eleventh International Conference on Learning Representations}, 2023.

\bibitem[Liu et~al.(2022{\natexlab{a}})Liu, Zheng, Su, Hu, Tan, and Chen]{liu2022exploiting}
Weiming Liu, Xiaolin Zheng, Jiajie Su, Mengling Hu, Yanchao Tan, and Chaochao Chen.
\newblock Exploiting variational domain-invariant user embedding for partially overlapped cross domain recommendation.
\newblock In \emph{Proceedings of the 45th International ACM SIGIR conference on research and development in information retrieval}, pages 312--321, 2022{\natexlab{a}}.

\bibitem[Liu et~al.(2023{\natexlab{a}})Liu, Zheng, Chen, Su, Liao, Hu, and Tan]{liu2023joint}
Weiming Liu, Xiaolin Zheng, Chaochao Chen, Jiajie Su, Xinting Liao, Mengling Hu, and Yanchao Tan.
\newblock Joint internal multi-interest exploration and external domain alignment for cross domain sequential recommendation.
\newblock In \emph{Proceedings of the ACM web conference 2023}, pages 383--394, 2023{\natexlab{a}}.

\bibitem[Liu et~al.(2023{\natexlab{b}})Liu, Zheng, Su, Zheng, Chen, and Hu]{liu2023contrastive}
Weiming Liu, Xiaolin Zheng, Jiajie Su, Longfei Zheng, Chaochao Chen, and Mengling Hu.
\newblock Contrastive proxy kernel stein path alignment for cross-domain cold-start recommendation.
\newblock \emph{IEEE Transactions on Knowledge and Data Engineering}, 35\penalty0 (11):\penalty0 11216--11230, 2023{\natexlab{b}}.

\bibitem[Liu et~al.(2024{\natexlab{a}})Liu, Chen, Liao, Hu, Su, Tan, and Wang]{liu2024user}
Weiming Liu, Chaochao Chen, Xinting Liao, Mengling Hu, Jiajie Su, Yanchao Tan, and Fan Wang.
\newblock User distribution mapping modelling with collaborative filtering for cross domain recommendation.
\newblock In \emph{Proceedings of the ACM Web Conference 2024}, pages 334--343, 2024{\natexlab{a}}.

\bibitem[Liu et~al.(2024{\natexlab{b}})Liu, Chen, Liao, Hu, Tan, Wang, Zheng, and Ong]{liu2024learning}
Weiming Liu, Chaochao Chen, Xinting Liao, Mengling Hu, Yanchao Tan, Fan Wang, Xiaolin Zheng, and Yew~Soon Ong.
\newblock Learning accurate and bidirectional transformation via dynamic embedding transportation for cross-domain recommendation.
\newblock In \emph{Proceedings of the AAAI Conference on Artificial Intelligence}, pages 8815--8823, 2024{\natexlab{b}}.

\bibitem[Liu et~al.(2024{\natexlab{c}})Liu, Zheng, Chen, Xu, Liao, Wang, Tan, and Ong]{liu2024reducing}
Weiming Liu, Xiaolin Zheng, Chaochao Chen, Jiahe Xu, Xinting Liao, Fan Wang, Yanchao Tan, and Yew-Soon Ong.
\newblock Reducing item discrepancy via differentially private robust embedding alignment for privacy-preserving cross domain recommendation.
\newblock In \emph{Forty-first International Conference on Machine Learning}, 2024{\natexlab{c}}.

\bibitem[Liu et~al.(2022{\natexlab{b}})Liu, Gong, and Liu]{liu2022flow}
Xingchao Liu, Chengyue Gong, and Qiang Liu.
\newblock Flow straight and fast: Learning to generate and transfer data with rectified flow.
\newblock \emph{arXiv preprint arXiv:2209.03003}, 2022{\natexlab{b}}.

\bibitem[Liu et~al.(2022{\natexlab{c}})Liu, Wu, Zhou, Du, Wu, Lin, and Liu]{liu2022audio}
Xian Liu, Qianyi Wu, Hang Zhou, Yuanqi Du, Wayne Wu, Dahua Lin, and Ziwei Liu.
\newblock Audio-driven co-speech gesture video generation.
\newblock \emph{Advances in Neural Information Processing Systems}, 35:\penalty0 21386--21399, 2022{\natexlab{c}}.

\bibitem[Mahmood et~al.(2019)Mahmood, Ghorbani, Troje, Pons-Moll, and Black]{mahmood2019amass}
Naureen Mahmood, Nima Ghorbani, Nikolaus~F Troje, Gerard Pons-Moll, and Michael~J Black.
\newblock Amass: Archive of motion capture as surface shapes.
\newblock In \emph{Proceedings of the IEEE/CVF international conference on computer vision}, pages 5442--5451, 2019.

\bibitem[Mittal et~al.(2021)Mittal, Engel, Hawthorne, and Simon]{mittal2021symbolicdiffusion}
Gautam Mittal, Jesse Engel, Curtis Hawthorne, and Ian Simon.
\newblock Symbolic music generation with diffusion models.
\newblock In \emph{Proceedings of the 22nd International Society for Music Information Retrieval Conference}, 2021.

\bibitem[Odena et~al.(2017)]{odena2017conditional}
Odena et~al.
\newblock Conditional image synthesis with auxiliary classifier gans.
\newblock In \emph{International conference on machine learning}, pages 2642--2651. PMLR, 2017.

\bibitem[Petrovich et~al.(2021)Petrovich, Black, and Varol]{petrovich2021action}
Mathis Petrovich, Michael~J Black, and G{\"u}l Varol.
\newblock Action-conditioned 3d human motion synthesis with transformer vae.
\newblock In \emph{Proceedings of the IEEE/CVF International Conference on Computer Vision}, pages 10985--10995, 2021.

\bibitem[Petrovich et~al.(2022)Petrovich, Black, and Varol]{petrovich2022temos}
Mathis Petrovich, Michael~J Black, and G{\"u}l Varol.
\newblock Temos: Generating diverse human motions from textual descriptions.
\newblock In \emph{European Conference on Computer Vision}, pages 480--497. Springer, 2022.

\bibitem[Peyr{\'e} et~al.(2019)Peyr{\'e}, Cuturi, et~al.]{peyre2019computational}
Gabriel Peyr{\'e}, Marco Cuturi, et~al.
\newblock Computational optimal transport: With applications to data science.
\newblock \emph{Foundations and Trends{\textregistered} in Machine Learning}, 11\penalty0 (5-6):\penalty0 355--607, 2019.

\bibitem[Raab et~al.(2023)Raab, Leibovitch, Li, Aberman, Sorkine-Hornung, and Cohen-Or]{raab2023modi}
Sigal Raab, Inbal Leibovitch, Peizhuo Li, Kfir Aberman, Olga Sorkine-Hornung, and Daniel Cohen-Or.
\newblock Modi: Unconditional motion synthesis from diverse data.
\newblock In \emph{Proceedings of the IEEE/CVF Conference on Computer Vision and Pattern Recognition}, pages 13873--13883, 2023.

\bibitem[Sajjadi et~al.(2018)Sajjadi, Bachem, Lucic, Bousquet, and Gelly]{sajjadi2018assessing}
Mehdi~SM Sajjadi, Olivier Bachem, Mario Lucic, Olivier Bousquet, and Sylvain Gelly.
\newblock Assessing generative models via precision and recall.
\newblock \emph{Advances in neural information processing systems}, 31, 2018.

\bibitem[Song et~al.(2020)Song, Meng, and Ermon]{song2020denoising}
Jiaming Song, Chenlin Meng, and Stefano Ermon.
\newblock Denoising diffusion implicit models.
\newblock \emph{arXiv preprint arXiv:2010.02502}, 2020.

\bibitem[Song and Ermon(2020)]{song2020improved}
Yang Song and Stefano Ermon.
\newblock Improved techniques for training score-based generative models.
\newblock \emph{Advances in neural information processing systems}, 33:\penalty0 12438--12448, 2020.

\bibitem[Song et~al.(2021)Song, Sohl-Dickstein, Kingma, Kumar, Ermon, and Poole]{song_score-based_2021}
Yang Song, Jascha Sohl-Dickstein, Diederik~P Kingma, Abhishek Kumar, Stefano Ermon, and Ben Poole.
\newblock {SCORE}-{BASED} {GENERATIVE} {MODELING} {THROUGH} {STOCHASTIC} {DIFFERENTIAL} {EQUATIONS}.
\newblock 2021.

\bibitem[Song et~al.(2023)Song, Dhariwal, Chen, and Sutskever]{song2023consistency}
Yang Song, Prafulla Dhariwal, Mark Chen, and Ilya Sutskever.
\newblock Consistency models.
\newblock \emph{arXiv preprint arXiv:2303.01469}, 2023.

\bibitem[Sun et~al.(2021)Sun, Lin, Han, Hu, Xu, and Zheng]{sun2021action}
Jiangxin Sun, Zihang Lin, Xintong Han, Jian-Fang Hu, Jia Xu, and Wei-Shi Zheng.
\newblock Action-guided 3d human motion prediction.
\newblock \emph{Advances in Neural Information Processing Systems}, 34:\penalty0 30169--30180, 2021.

\bibitem[Sun et~al.(2023)Sun, Chen, Yang, and Nishida]{NEURIPS2023_4c9477b9}
Zitang Sun, Yen-Ju Chen, Yung-Hao Yang, and Shin\textquotesingle~ya Nishida.
\newblock Modeling human visual motion processing with trainable motion energy sensing and a self-attention network.
\newblock In \emph{Advances in Neural Information Processing Systems}, pages 24335--24348. Curran Associates, Inc., 2023.

\bibitem[Tan et~al.(2023)Tan, Nam, Nam, and Noh]{10.1145/3610543.3626164}
Vanessa Tan, Junghyun Nam, Juhan Nam, and Junyong Noh.
\newblock Motion to dance music generation using latent diffusion model.
\newblock In \emph{SIGGRAPH Asia 2023 Technical Communications}, New York, NY, USA, 2023. Association for Computing Machinery.

\bibitem[Tang et~al.(2024)Tang, Sun, Lin, Zheng, Hu, et~al.]{tang2024temporal}
Jianwei Tang, Jiangxin Sun, Xiaotong Lin, Wei-Shi Zheng, Jian-Fang Hu, et~al.
\newblock Temporal continual learning with prior compensation for human motion prediction.
\newblock \emph{Advances in Neural Information Processing Systems}, 36, 2024.

\bibitem[Tevet et~al.(2022)Tevet, Raab, Gordon, Shafir, Cohen-Or, and Bermano]{tevet_human_2022}
Guy Tevet, Sigal Raab, Brian Gordon, Yonatan Shafir, Daniel Cohen-Or, and Amit~H. Bermano.
\newblock Human {Motion} {Diffusion} {Model}, 2022.
\newblock arXiv:2209.14916 [cs].

\bibitem[Vahdat et~al.(2021)Vahdat, Kreis, and Kautz]{vahdat2021score}
Arash Vahdat, Karsten Kreis, and Jan Kautz.
\newblock Score-based generative modeling in latent space.
\newblock \emph{Advances in neural information processing systems}, 34:\penalty0 11287--11302, 2021.

\bibitem[Van Den~Oord et~al.(2017)Van Den~Oord, Vinyals, et~al.]{van2017neural}
Aaron Van Den~Oord, Oriol Vinyals, et~al.
\newblock Neural discrete representation learning.
\newblock \emph{Advances in neural information processing systems}, 30, 2017.

\bibitem[Vaswani et~al.(2017)Vaswani, Shazeer, Parmar, Uszkoreit, Jones, Gomez, Kaiser, and Polosukhin]{vaswani2017attention}
Ashish Vaswani, Noam Shazeer, Niki Parmar, Jakob Uszkoreit, Llion Jones, Aidan~N Gomez, {\L}ukasz Kaiser, and Illia Polosukhin.
\newblock Attention is all you need.
\newblock \emph{Advances in neural information processing systems}, 30, 2017.

\bibitem[Wang et~al.(2024)Wang, Li, and Cui]{wang2024incomplete}
Yuanzhi Wang, Yong Li, and Zhen Cui.
\newblock Incomplete multimodality-diffused emotion recognition.
\newblock \emph{Advances in Neural Information Processing Systems}, 36, 2024.

\bibitem[Wang et~al.(2022)Wang, Chen, Liu, Zhu, Liang, and Huang]{NEURIPS2022_6030db51}
Zan Wang, Yixin Chen, Tengyu Liu, Yixin Zhu, Wei Liang, and Siyuan Huang.
\newblock Humanise: Language-conditioned human motion generation in 3d scenes.
\newblock In \emph{Advances in Neural Information Processing Systems}, pages 14959--14971. Curran Associates, Inc., 2022.

\bibitem[Wei et~al.(2023)Wei, Sun, Li, Lu, Li, Sun, and Hu]{wei2023human}
Dong Wei, Huaijiang Sun, Bin Li, Jianfeng Lu, Weiqing Li, Xiaoning Sun, and Shengxiang Hu.
\newblock Human joint kinematics diffusion-refinement for stochastic motion prediction.
\newblock In \emph{Proceedings of the AAAI Conference on Artificial Intelligence}, pages 6110--6118, 2023.

\bibitem[Wu et~al.(2023)Wu, Wang, Gong, Liu, Xiong, Ranjan, Krishnamoorthi, Chandra, and Liu]{wu2023fast}
Lemeng Wu, Dilin Wang, Chengyue Gong, Xingchao Liu, Yunyang Xiong, Rakesh Ranjan, Raghuraman Krishnamoorthi, Vikas Chandra, and Qiang Liu.
\newblock Fast point cloud generation with straight flows.
\newblock In \emph{Proceedings of the IEEE/CVF conference on computer vision and pattern recognition}, pages 9445--9454, 2023.

\bibitem[Yan et~al.(2018)Yan, Rastogi, Villegas, Sunkavalli, Shechtman, Hadap, Yumer, and Lee]{yan2018mt}
Xinchen Yan, Akash Rastogi, Ruben Villegas, Kalyan Sunkavalli, Eli Shechtman, Sunil Hadap, Ersin Yumer, and Honglak Lee.
\newblock Mt-vae: Learning motion transformations to generate multimodal human dynamics.
\newblock In \emph{Proceedings of the European conference on computer vision (ECCV)}, pages 265--281, 2018.

\bibitem[Yu et~al.(2023)Yu, Fan, Hou, Pei, Ge, Yang, Zhou, Zhang, and Zhang]{yu3}
Hua Yu, Xuanzhe Fan, Yaqing Hou, Wenbin Pei, Hongwei Ge, Xin Yang, Dongsheng Zhou, Qiang Zhang, and Mengjie Zhang.
\newblock Toward realistic 3d human motion prediction with a spatio-temporal cross- transformer approach.
\newblock \emph{IEEE Transactions on Circuits and Systems for Video Technology}, 33\penalty0 (10):\penalty0 5707--5720, 2023.

\bibitem[Yu et~al.(2024{\natexlab{a}})Yu, Hou, Pei, Ong, and Zhang]{yu1}
Hua Yu, Yaqing Hou, Wenbin Pei, Yew-Soon Ong, and Qiang Zhang.
\newblock Divdiff: A conditional diffusion model for diverse human motion prediction.
\newblock \emph{IEEE Transactions on Multimedia}, pages 1--12, 2024{\natexlab{a}}.

\bibitem[Yu et~al.(2024{\natexlab{b}})Yu, Liu, Bai, Gui, Hou, Ong, and Zhang]{yu2}
Hua Yu, Weiming Liu, Jiapeng Bai, Xu Gui, Yaqing Hou, YewSoon Ong, and Qiang Zhang.
\newblock Towards efficient and diverse generative model for unconditional human motion synthesis.
\newblock In \emph{Proceedings of the 32nd ACM International Conference on Multimedia}, page 2535–2544, New York, NY, USA, 2024{\natexlab{b}}. Association for Computing Machinery.

\bibitem[Yuan et~al.(2020)]{yuan2020dlow}
Yuan et~al.
\newblock Dlow: Diversifying latent flows for diverse human motion prediction.
\newblock In \emph{Computer Vision--ECCV 2020: 16th European Conference, Glasgow, UK, August 23--28, 2020, Proceedings, Part IX 16}, pages 346--364. Springer, 2020.

\bibitem[Yuan et~al.(2023)Yuan, Song, Iqbal, Vahdat, and Kautz]{yuan2023physdiff}
Ye Yuan, Jiaming Song, Umar Iqbal, Arash Vahdat, and Jan Kautz.
\newblock Physdiff: Physics-guided human motion diffusion model.
\newblock In \emph{Proceedings of the IEEE/CVF International Conference on Computer Vision}, pages 16010--16021, 2023.

\bibitem[Zhang et~al.(2023)Zhang, Zhang, Cun, Zhang, Zhao, Lu, Shen, and Shan]{Zhang_2023_CVPR}
Jianrong Zhang, Yangsong Zhang, Xiaodong Cun, Yong Zhang, Hongwei Zhao, Hongtao Lu, Xi Shen, and Ying Shan.
\newblock Generating human motion from textual descriptions with discrete representations.
\newblock In \emph{Proceedings of the IEEE/CVF Conference on Computer Vision and Pattern Recognition (CVPR)}, pages 14730--14740, 2023.

\bibitem[Zhang et~al.(2021)Zhang, Black, and Tang]{zhang2021we}
Yan Zhang, Michael~J Black, and Siyu Tang.
\newblock We are more than our joints: Predicting how 3d bodies move.
\newblock In \emph{Proceedings of the IEEE/CVF Conference on Computer Vision and Pattern Recognition}, pages 3372--3382, 2021.

\bibitem[Zhang et~al.(2024)Zhang, Huang, Liu, Tang, Lu, Chen, Bai, Chu, Yu, and Ouyang]{zhang2024motiongpt}
Yaqi Zhang, Di Huang, Bin Liu, Shixiang Tang, Yan Lu, Lu Chen, Lei Bai, Qi Chu, Nenghai Yu, and Wanli Ouyang.
\newblock Motiongpt: Finetuned llms are general-purpose motion generators.
\newblock In \emph{Proceedings of the AAAI Conference on Artificial Intelligence}, pages 7368--7376, 2024.

\end{thebibliography}
